\pdfoutput=1
\documentclass{article}

\usepackage[preprint]{nmod_2023}

\usepackage[utf8]{inputenc}
\usepackage[T1]{fontenc}
\usepackage{url}
\usepackage{booktabs}
\usepackage{amsfonts}
\usepackage{nicefrac}
\usepackage{microtype}
\usepackage{amsmath,amssymb,amsthm}
\usepackage{verbatim}
\usepackage{graphics}
\usepackage{epsfig}
\usepackage{bm}
\usepackage[dvipsnames]{xcolor}
\usepackage{lipsum}

\newtheorem{theorem}{Theorem}
\newtheorem{example}{Example}
\newtheorem{assumption}{Assumption}

\newcommand\blfootnote[1]{%
  \begingroup
  \renewcommand\thefootnote{}\footnote{#1}%
  \addtocounter{footnote}{-1}%
  \endgroup
}

\newcommand{\indep}{\perp\!\!\!\!\perp}

\title{Learning under random distributional shifts}

\author{
  Kirk Bansak$^*$ \\
  Department of Political Science \\
  Univeristy of California, Berkeley \\
  \texttt{kbansak@berkeley.edu} \\
   \And
   Elisabeth Paulson$^*$ \\
   Technology and Operations Management Unit \\
   Harvard Business School \\
   \texttt{epaulson@hbs.edu} \\
   \And
   Dominik Rothenhäusler$^*$ \\
   Department of Statistics \\
   Stanford University \\
   \texttt{rdominik@stanford.edu} \\
}

\begin{document}

\maketitle

\begin{abstract}
Many existing approaches for generating predictions in settings with distribution shift model distribution shifts as adversarial or low-rank in suitable representations. In various real-world settings, however, we might expect shifts to arise through the superposition of many small and random changes in the population and environment. Thus, we consider a class of random distribution shift models that capture arbitrary changes in the underlying covariate space, and dense, random shocks to the relationship between the covariates and the outcomes. In this setting, we characterize the benefits and drawbacks of several alternative prediction strategies: the \emph{standard approach} that directly predicts the long-term outcomes of interest, the \emph{proxy approach} that directly predicts a shorter-term proxy outcome, and a \emph{hybrid approach} that utilizes both the long-term policy outcome and (shorter-term) proxy outcome(s). We show that the hybrid approach is robust to the strength of the distribution shift and the proxy relationship. We apply this method to datasets in two high-impact domains: asylum-seeker resettlement and early childhood education. In both settings, we find that the proposed approach results in substantially lower mean-squared error than current approaches. \blfootnote{$^*$Faculty Affiliate, Immigration Policy Lab, Stanford University and ETH Zurich.}
\end{abstract}

\section{INTRODUCTION}
Distribution shift---changes in the underlying data distribution over time or across locations---is a persistent obstacle to generating high-quality, long-term predictions in various real-world settings. For example, consider the burgeoning research on the algorithmic assignment of refugees and asylum seekers to geographic locations within host countries. This approach, introduced by \cite{bansak2018improving}, uses machine learning models trained on historical data to generate counterfactual outcome predictions for every refugee--location combination upon arrival. Then, these predictions are used to assign the refugee and/or aslyum seeker family to a particular location. This approach has ongoing implementations in Switzerland, the Netherlands, and the US.

However, the efficacy of these implementations largely depends on the accuracy of the underlying outcome predictions. One persistent challenge in generating these predictions is the inherent nonstationarity across time and locations, which leads to a drop in performance between the training and the deployment environment. This problem is further compounded by the fact that policy outcomes of interest are often long term. In the case of refugee and asylum seeker resettlement, longer-term outcomes better capture the experience and welfare of refugees than short-term outcomes that may reflect transient dynamics. This problem is not unique to the refugee or asylum seeker resettlement domain; rather, decision-making in many high-impact settings requires making long-term predictions in changing environments. 

As a result, there has been a surge of interest in robust and generalizable machine learning. Distribution shift models can generally be grouped into \emph{adversarial} and \emph{representation}-based approaches. The adversarial approach focuses on worst-case perturbations that either appear in the training or the deployment environment. 
The goal of representation-based approaches is to decompose the data into spurious and invariant components, and then to develop a prediction method that leverages the invariances in the data. 

Both approaches have shown some success, but do not consistently beat empirical risk minimization \citep{gulrajani2020search,koh2021wilds}. There might be several contributing factors. First, it is challenging to tune models under distributional shifts. In addition, adversarial-based approaches can be conservative for real-world settings, such as in the example above. Representation-based approaches rely on an assumption that the change in distribution can be represented in low-dimension. This might be unrealistic for distributional changes that arise through the superposition of thousands (or millions) of small changes in circumstances.

In contrast to these settings, we are interested in generating predictions amidst complex social and economic phenomena where there are no adversarial agents, but rather dense and inherently unpredictable shifts driven by a confluence of broader social, economic, and natural forces. As put by Thorsten Drautzberg, a senior economist at the Federal Reserve of Philadelphia, mainstream economics sees business cycles as driven by the `random summation of random causes' \citep{drautzburg2019}. 

As a result, we consider a class of distribution shifts that capture changes that may arise through the superposition of many random changes. Models for random distributions are popular in Bayesian nonparametrics \citep{ghosal2017fundamentals}, but only recently have been used to model distribution shift in frequentist estimation problems \citep{jeong2022calibrated,rothenhausler2022distributionally}. In this work, we consider the problem of making robust \emph{predictions} under random distribution shifts in a frequentist setting.

This paper formalizes and characterizes the benefits and drawbacks of several alternative prediction strategies in settings with random distribution shift: the \emph{standard approach} that directly predicts the long-term outcomes of interest, the \emph{proxy approach} that directly predicts a shorter-term proxy outcome, and a \emph{hybrid approach} that utilizes both the long-term policy outcome and (shorter-term) proxy outcome(s). Thus, we address the following question: Given the goal of maximizing a specific policy outcome, how should we leverage proxy outcomes and policy outcomes in a randomly changing environment?

\subsection{Contributions}

This study makes the following contributions:

\noindent \textbf{Methodological}. We contribute to the research agenda on prediction under distribution shifts (see related work below) by providing the first comparison of models under random distribution shifts in a nonparametric setting. Theorem \ref{thm:mse} provides an exact characterization of the weaknesses of each approach of three approaches: the standard approach is ineffective when distribution shift is large, and the proxy approach falters if the proxy relationship is not strong enough. The hybrid approach, on the other hand, is robust to both of these failure points.

\noindent\textbf{Empirical}. We establish that the robustness of the hybrid approach is not simply a theoretical novelty but a matter of real-world, practical significance by presenting empirical evidence from two high-impact domains: geographic assignment of asylum seekers in the Netherlands and an early childhood education intervention in the US. In both cases, the proposed approach results in substantially better predictions of the relevant long-term policy outcome than the standard or proxy approach.

\section{RELATED WORK}

\noindent\textbf{Self-training.} A popular method for combining labeled data with unlabeled data is self-training \citep{scudder1965probability,semisup}. In its most basic form, one starts by learning a prediction mechanism only for the labeled data. Then, based on imputations on the unlabeled data, the prediction mechanism is re-trained on its own predictions for the unlabeled data and the original labeled data. In the past few years, there has been an increasing interest in studying the theoretical properties of self-training, in particular, its robustness under distribution shift \citep{carmon2019unlabeled,chen2020self,raghunathan2020understanding,kumar2020understanding}. In contrast to self-training, due to the distribution shift structure, we discard the outdated data during the re-training stage.

\noindent\textbf{Pre-training.} In the transfer learning literature, it is common to learn a feature representation by regressing auxiliary data on the covariates of a large data set and then using this as a feature vector on the smaller (target) data set, either by updating the feature vector or by regressing the final outcome on the feature representation
\citep{caruana1997multitask,weiss2016survey,hendrycks2019using}. Overall, this has some similarities with the two-step procedure $\tau^C$. Practically, one difference is that in the second stage, we do not observe the outcome, so we regress imputed outcomes on the new observed data. Another difference is that in pre-training one often assumes that the final prediction model is within a neighborhood of the pre-trained prediction model, with a low-dimensional update to the parameter. In our model, the prediction model may change in a random and dense fashion, driven by a confluence of social, economic, and natural forces.

\noindent\textbf{Adversarial machine learning.} There is a rich literature that models distributional perturbations as adversarial \citep{huber1964robust,ben2009robust,maronna2019robust,biggio2018wild}. The intuition is that a small change in input at the training or prediction stage should not change the output. Adversarial inputs are somewhat pessimistic for our problem setup, where the changes occur due to natural shifts over time (as discussed above).

\noindent\textbf{Invariance-based approaches.} Instead of considering small shifts, another line of work considers distribution shifts that lie on subspaces and relies on the assumption that certain conditional probabilities or representations stay invariant across settings. In other words, these works study representations and procedures that use invariances to transfer across settings \citep[e.g.,][]{ganin2015unsupervised,rojas2018invariant,arjovsky2019invariant,rothenhausler2021anchor}. For an overview, see \cite{chen2021domain}. These methods have shown some success, but do not consistently outperform empirical risk minimization \citep{gulrajani2020search,koh2021wilds}. One major difference, as described above, is that in our setting we expect changes to arise through a superposition of many random environmental and economic changes, which violates the invariance assumption.

\noindent \textbf{Surrogate outcomes}. When policy outcomes are long-term, surrogate outcomes are a common tool to guide policy decisions on a faster time scale \citep{prentice1989surrogate}. In the traditional setting, the policy maker has access to historical data containing the long-term outcome but not the treatment decision, and experimental data containing the treatment decision but not the long-term outcome \citep{yang2020targeting, athey2019surrogate}. Thus, the problem can be viewed as a missing data issue. By contrast, in this paper the historical data contains both treatment decisions and the long-term outcome of interest, and the motivation for using a shorter-term outcome is distribution shift. 

\section{SETUP}

\subsection{Preliminaries}\label{sec:prelims}

Consider a finite time horizon with three distinct periods: period $0$ (the present), period $-1$ (one period prior), and period $-2$ (two periods prior). In each period, a large\footnote{For most of the paper, we make an infinite-data assumption. In Section \ref{sec:finite} we discuss finite-sample considerations.} sample of i.i.d. units (e.g., refugees) arrive with some vector of background characteristics, $X=x$, and their outcomes are observed for at least two subsequent periods. Let $Y_2$---the outcome observed after two periods---be the policy outcome of interest, and let $Y_1$ be a related but shorter-term outcome measured one period after arrival\footnote{Appendix \ref{app:extensions} discusses how to extend the proposed methods to more than two time periods and outcomes}.

For units who arrive in any period $t$ and the variables defined above, we posit the existence of a tuple generated according to a probability distribution $P_t$:
\begin{equation*}
(Y_1,Y_2,X)_{t} \sim P_t.
\end{equation*}
This paper is focused on predicting $Y_2$ for units who arrive in the present period, that is $ \tau(x) := \mathbb{E}_0[Y_2 | X=x]$. 

The problem is that we do not observe $Y_2$ (nor do we observe $Y_1$) for the cohort that arrives in period $0$ at the time of prediction. Thus, we do not have data to directly fit the target $\mathbb{E}_0[Y_2|X=x]$. Instead, we observe the data $(Y_2, Y_1, X)_{-2} \sim P_{-2}$, $(Y_1, X)_{-1} \sim P_{-1}$, and $(X)_{0} \sim P_{0}$.

\subsection{Alternative approaches}

In the presence of distribution shift, there are (at least) three strategies one could pursue:
\begin{enumerate}
    \item $\tau^A(x) = \mathbb{E}_{-2}[Y_2 | X=x]$
    \item $ \tau^B(x) = \mathbb{E}_{-1}[ Y_1 | X=x]$
    \item $ \tau^C(x) = \mathbb{E}_{-1}[\mathbb{E}_{-2}[Y_2 | Y_1, X] | X=x]$
\end{enumerate}

The first approach estimates the policy outcome using the data from period $-2$.\footnote{In practice, data from earlier periods could also be used. This is also true for the second and third approaches.} In many settings such as refugee resettlement, this is the default approach in the literature. Unfortunately, its performance may suffer under distribution shift.

The second approach estimates the shorter-term outcome, $Y_1$, using data from period $-1$.
Relative to $\tau^A(x)$, the advantage of this strategy is that it employs more recent data, and hence is less susceptible to the effects of distribution shift. 
However, the performance of this approach hinges on the strength of the proxy outcome.

The third approach again estimates $Y_2$, but uses the data from both periods $-2$ and $-1$. The general idea behind the third strategy is that we want to devise the best available approximation of the density $P_0(y_2,y_1,x) = P_0(y_2 |y_1,x) P_0(y_1,x)$. We cannot know $P_0(y_2 |y_1,x)$, so we use the best guess $P_{-2}(y_2 |y_1,x)$. And we cannot know $P_0(y_1,x)$ so we use the best guess $P_{-1}(y_1,x)$, since $P_{-1}$ is likely closer to $P_0$ than is $P_{-2}$. Putting things together,
\begin{align*}
   \tau(x) &= \mathbb{E}_0[Y_2|X=x]= \mathbb{E}_0[\mathbb{E}_0[Y_2|X,Y_1]|X=x] \\
   & \approx \mathbb{E}_{-1}[\mathbb{E}_{-2}[Y_2|X,Y_1]|X=x]= \tau^C(x).
\end{align*}
In words, this strategy first regresses $Y_2$ on $X$ and $Y_1$ using the period $-2$ data, and then regresses that fitted model on $X$ using the period $-1$ data.

\subsection{Robust predictions under distribution shift.} 

Before diving into the theoretical model, let us informally discuss a robustness property of $\tau^C$. Namely, that $\tau^C$ can be seen as an interpolation estimator that behaves similarly to $\tau^A$ and ${\tau}^B$ in extreme cases. 

First, note that intuitively $\tau^B$ is problematic if $Y_1$ is not predictive of $Y_2$. In this case, $\tau^C(x) \approx \tau^A(x)$.
\begin{example}[Proxy violated]
Let us consider the case where  $Y_1$ is not predictive of $Y_2$, that is $\mathbb{E}_{-2}[Y_2|Y_1,X] \approx \mathbb{E}_{-2}[Y_2|X]$. Using the tower property,
\begin{align*}
    \tau^C (x) &= \mathbb{E}_{-1}[\mathbb{E}_{-2}[Y_2 | Y_1, X] | X=x]  \\
    &\approx \mathbb{E}_{-1}[\mathbb{E}_{-2}[Y_2 | X] | X=x]  \\
    &= \mathbb{E}_{-2}[Y_2|X=x]\\
    &= \tau^A(x).
\end{align*}
 \end{example}
 Now let us turn to a different extreme case. If $Y_1 \approx Y_2$, then one might prefer $\tau^B$ over $\tau^A$, since it uses more recent data. In this case, $\tau^C(x) \approx \tau^B(x)$.
 \begin{example}[$Y_1 \approx Y_2$]
   If $Y_1 \approx Y_2$, then $\mathbb{E}_{-2}[Y_2|Y_1,X] \approx Y_1$. Thus,
\begin{align*}
     \tau^B(x) &= \mathbb{E}_{-1}[ Y_1 | X=x] \\
    &\approx \mathbb{E}_{-1}[ \mathbb{E}_{-2}[Y_2|Y_1,X]| X=x] \\
    &= \tau^C(x).
\end{align*}
\end{example}
We see this interpolation property as an advantage of $\tau^C$. However, these extreme cases are not very realistic in practice. Thus, in the following, we will study settings in between these extreme cases. 

\section{PREDICTIONS UNDER DISTRIBUTION SHIFT}\label{sec:theory}

As motivated above, we will model distributional changes as random. To avoid measurability issues and simplify the discussion, we will focus on discrete distributions, that is, the random variable takes value in a finite alphabet $(X,Y_1,Y_2) \in \mathcal{X} \times \mathcal{Y}_1  \times \mathcal{Y}_2$. Note that this focus does not impose a serious practical constraint, as any continuous variables of interest can be arbitrarily coarsened to meet this requirement. We consider the distribution $P_{-2}(y_1,y_2|x)$ as fixed, and assume that $P_{-1}(y_1,y_2|x)$ differs randomly from $P_{-2}(y_1,y_2|x)$ and that $P_0(y_1,y_2|x)$ differs randomly from $P_{-1}(y_1,y_2|x)$. To be more specific, for $t \in \{ -2, -1 \} $ we write
\begin{equation*}
  S_{t}(y_1,y_2|x) =  P_{t+1}(y_1,y_2|x) - P_{t}(y_1,y_2|x),
\end{equation*}
where we call $S_{t}(y_1,y_2|x)$ a distributional shift variable. That is, between time $0$ and $-1$ and between time $-1$ and $-2$ the event probabilities get shifted by a random amount that can depend on $y_1$, $y_2$, and $x$. There are two constraints on the shift variables to make $P_{-1}$ and $P_{-2}$ well-defined probability measures. First, the shift variables must satisfy  $ 0 \le  P_{t}(y_1,y_2|x) + S_{t}(y_1,y_2|x)  \le 1.$ Furthermore, since
 $\sum_{y_1,y_2} P_{0}(y_1,y_2|x) = 1 = \sum_{y_1,y_2} P_{-1}(y_1,y_2|x)$ the shift variable $S_t$ has to satisfy $  \sum_{y_1,y_2} S_{t}(y_1,y_2|x) = 0$.

We allow the distribution to shift arbitrarily in $X$; however, to avoid issues of identifiability, we require the support of $X$ to be time invariant. Note that we also allow the shifts to be correlated across different $x$, that is we allow 
\begin{equation*}
\text{Cov}(S_t(y_1,y_2|x),S_t(y_1,y_2|x')) \neq 0
\end{equation*}
for $x \neq x'$. This allows for shifts that affect many units in a similar way (for example, an economic boom). We will assume that the perturbation process has (conditional) mean zero.
\begin{assumption}[Centered shifts]\label{ass:centered}
   $\mathbb{E}[ S_{-2}] = 0$ and $\mathbb{E}[ S_{-1} | S_{-2}] = 0$.
\end{assumption}
Intuitively, this means that the perturbation has no momentum: the perturbation that shifts $P_{-2}$ to $P_{-1}$ is uncorrelated with the perturbation that shifts $P_{-1}$ to $P_{0}$. This does not rule out memory/momentum driven by systematic factors (e.g. systematic or secular changes in the labor market) that can be captured by time and other covariates contained in $X$.

To be able to give an interpretable decomposition of the mean-squared error for different procedures, we put an additional working assumption on the distribution shift (which will be relaxed later). In Bayesian statistics, distributions are often modelled as random, with the Dirichlet process the ``default prior on spaces of probability measures'' \citep[Chapter 4]{ghosal2017fundamentals}. Motivated by this, we use a frequentist variant to describe how distributions shift across time. This assumption is mostly for interpretability and will be relaxed in Section~\ref{sec:asymmetric}. 
\begin{assumption}[Symmetry]\label{ass:scaling}
For some potentially unknown $\kappa_t$ with $0 < \kappa_t < 1$, for all events $\bullet \subseteq \mathcal{Y}_1 \times \mathcal{Y}_2$,
\begin{align*}
\begin{split}
        \text{Var}(S_{-2}(\bullet|x) ) &= \kappa_{-2} P_{-2}(\bullet|x)(1-P_{-2}(\bullet|x)), \\
    \text{Var}(S_{-1}(\bullet|x) | S_{-2}) &= \kappa_{-1} P_{-1}(\bullet|x)(1-P_{-1}(\bullet|x)).
\end{split}
\end{align*}
\end{assumption}
In Assumption~\ref{ass:scaling}, the event $\bullet$ and the variable $x$ are considered fixed, and the variance is over the randomness in the shift $S_{-2}$ (top equation) or $S_{-1}$ (bottom equation). We will not discuss how to construct such distribution shifts, but refer to reader to \cite{ghosal2017fundamentals} and \cite{jeong2022calibrated}.

Intuitively, this perturbation process captures ``well-behaved'' or ``symmetric'' distribution shift \citep{jeong2022calibrated}, where the perturbation of an event depends on its initial probability: events that have a small probability are only perturbed by a small amount. One can think about the unknown scaling factor $\kappa_t \in (0,1)$ as a measure of the strength of distribution shift between $P_{t+1}$ and $P_{t}$. 

We now present a theorem that compares the predictive performance of the three approaches under distribution shift, in the infinite-data case.  Finite-sample considerations are discussed in Section~\ref{sec:finite}.  The proof can be found in Appendix \ref{sec:proofs}. For clarity, we will assume that $Y_1$ and $Y_2$ are on the same scale. That is, a positive linear transformation has been applied to $Y_1$ in a pre-processing step such that $\mathbb{E}_{-2}[(\mathbb{E}_{-2}[Y_1|X]-  \mathbb{E}_{-2}[Y_2|X])^2]$ is as low as possible. Strictly speaking, the mathematical results below do not require this scaling, but the interpretation is clearer with scaling. When $Y_1$ is rescaled, for clarity we will rename $\tau^B$ as $\tilde{\tau}^B$ and $Y_1$ as $\tilde Y_1$.

\begin{theorem}[Comparison under symmetric shifts]\label{thm:mse}
Under Assumption~\ref{ass:scaling} and Assumption~\ref{ass:centered},
\begin{align*}
  \mathrm{MSE}( \tau^C) =& \underbrace{ \kappa_{-2} \mathbb{E}_{-2}[( Y_2 - \mathbb{E}_{-2}[Y_2|Y_1,X]  )^2]}_{\text{shift between $-2$ and $-1$}} \\
  &+ K_1 + O(\kappa_{-1} \kappa_{-2}) \\
   \mathrm{MSE}( \tau^A) =& \underbrace{ \kappa_{-2} \mathbb{E}_{-2}[(Y_2 - \mathbb{E}_{-2}[Y_2|X] )^2]}_{\text{shift between $-2$ and $-1$}} \\
   &+K_1  + O(\kappa_{-1} \kappa_{-2})  \\
    \mathrm{MSE}(\tilde{\tau}^B ) =& \underbrace{\mathbb{E}_{-2}[( \mathbb{E}_{-2}[Y_2 - \tilde Y_1|X])^2]}_{\text{error due to using proxy outcome}} \\
    &+  \underbrace{\kappa_{-2} \mathbb{E}_{-2}[(Y_2 - \tilde Y_1 - \mathbb{E}_{-2}[Y_2 - \tilde Y_1|X] )^2]}_{\text{inflation of proxy error under shift}} \\
    &+K_1 + O(\kappa_{-1} \kappa_{-2}) 
\end{align*}
where $K_1:=\kappa_{-1} \mathbb{E}_{-2}[(Y_2 - \mathbb{E}_{-2}[Y_2|X])^2]$ is the shift between $-1$ and $0$ that impacts all methods equally.
Here, $\text{MSE}(\tau^{\bullet}) = \mathbb{E}[(\tau^{\bullet}(X) - \tau(X))^2]$, where the outer expectation is both over the randomness in distribution shift and the randomness in $X$.
\end{theorem}
Let us discuss how to interpret the results for the case of $\tau^C$. First, since $\kappa_t \in (0,1)$, $\kappa_{-1} \kappa_{-2}$ is usually of lower magnitude than the other error terms. At a high level, one can think about the error as
\begin{equation*}
    \text{MSE}(\tau^C) = \sum_{\text{shifts}} \text{strength of shift} \cdot \text{ sensitivity }
\end{equation*}
One part of the error depends on the shift between $-2$ and $-1$, and one part depends on the shift between $-1$ and $0$. The strength of the shift enters the equation via $\kappa_{-1}$ and $\kappa_{-2}$, and is multiplied by the sensitivity of the $\tau^C(x)$ with respect to that particular distributional change. The sensitivity of $\tau^C(x)$ depends on the noise level $\mathbb{E}_{-2}[( Y_2 - \mathbb{E}_{-2}[Y_2|Y_1,X]  )^2]$. This makes sense intuitively: if the noise is high that means there are important unobserved variables besides $X$ that determine the value of $Y_2$. Since those unobserved  variables can shift in distribution over time, generalization may suffer. The estimator $\tau^A$ has a similar decomposition and interpretation as $\tau^C$.\ 

Now let us turn to the interpretation of the error of $\tilde{\tau}^B$. There are three terms. The first term captures whether $\tilde Y_1$ is a reasonable proxy for $Y_2$ in the absence of distribution shift. Even if $\tilde Y_1$ is a good proxy under $P_{-2}$, due to distribution shift this might change between $P_{-2}$ and $P_{-1}$. This is captured in the second term. The second term may be zero, for example in the case where $\tilde Y_1 \stackrel{\text{a.s.}}{=} Y_2$. Finally, the last term corresponds to the shift between $P_{-1}$ and $P_0$, which is the same for $\tau^C$, $\tau^A$, and $\tilde{\tau}^B$. It can be thought of as an irreducible error since all procedures are equally affected by the shift between $P_{-1}$ and $P_0$.

\subsection{Comparison of the three approaches}
 
In the following, we build further intuition for the strengths and weaknesses of the approaches.

\textbf{$\tau^A$ versus $\tau^C$.} On a high level, $\tau^A$ is more affected by shifts between $-2$ and $-1$ than $\tau^C$. In the theorem this is reflected by the fact that $\kappa_{-2}$ is multiplied by sensitivity factor $\mathbb{E}_{-2}[( Y_2 -  \mathbb{E}_{-2}[Y_2|Y_1,X])^2]$, which is always smaller, and can be much smaller, than $\tau^A$'s sensitivity factor $\mathbb{E}_{-2}[( Y_2 - \mathbb{E}_{-2}[Y_2|X]  )^2]$. Using the theorem,
\begin{equation*}
    \text{MSE}(\tau^C) \ll \text{MSE}(\tau^A) + O(\kappa_{-1} \kappa_{-2}).
\end{equation*}
By this argument, $\tau^C$ is generally preferable over $\tau^A$ (and is always preferable under the assumptions made above).
That being said, there may be real-world situations where one would prefer $\tau^A$ over $\tau^C$. For example, if $P_{-1}$ is the time period of a pandemic, then the optimal prediction mechanism might change drastically for this particular year ($P_{-1}$), and then switch back to the original mechanism, leading to $P_{-2} \approx P_0$. In this case, the (random) distribution shifts $S_{-1}$ and $S_{-2}$ are negatively correlated.

\textbf{$\tilde{\tau}^B$ versus $\tau^C$.} The comparison between $\tilde{\tau}^B$ and $\tau^C$ is slightly more technical. Both have the same dependence on $\kappa_{-1}$, the shift between $P_{0}$ and $P_{-1}$. However, they have a different sensitivity with respect to the shift between $P_{-2}$ and $P_{-1}$. This sensitivity is  lower for $\tau^C$:
\begin{align*}
    &\mathbb{E}_{-2}[(Y_2 - \mathbb{E}_{-2}[Y_2|Y_1,X])^2]  \\
    &= \min_{g(Y_1,X)} \mathbb{E}_{-2}[(Y_2 - g(Y_1,X))^2] \\
    &\le \mathbb{E}_{-2}[(Y_2 - \tilde Y_1 - \mathbb{E}_{-2}[Y_2 - \tilde Y_1|X] )^2].
\end{align*}
Thus, $\tau^C$ also outperforms $\tilde{\tau}^B$ (up to lower-order terms), and the gap will generally grow as the correlation of $Y_2$ and $\tilde Y_1$ decreases.

\subsection{Asymmetric shifts}\label{sec:asymmetric}

Theorem~\ref{thm:mse} provides interpretable error terms that show how different procedures behave under various type of shifts, in particular that $\tau^C$ is non-inferior to the other approaches. The conclusion that $\tau^C$ is non-inferior to the other approaches also holds under asymmetric shifts, if we assume that the distribution of $y_2$ given $y_1$ and $x$ stays invariant.
\begin{theorem}[Comparison under asymmetric shifts]\label{theorem:asymmetric}
    Under Assumption~\ref{ass:centered}, and assuming that $P_{-2}(y_2|y_1,x) = P_{-1}(y_2|y_1,x) = P_0(y_2|y_1,x)$,
    \begin{align*}
        \mathrm{MSE}( \tau^C) \le \mathrm{min}( \mathrm{MSE}( \tau^B), \mathrm{MSE}( \tau^A)).
    \end{align*}
\end{theorem}

The main intuition for this phenomenon is that the third strategy corresponds to a ``best guess'' under random centered shifts, as explained in Section \ref{sec:prelims}. The proof of Theorem \ref{theorem:asymmetric} is in Appendix \ref{sec:proofs}.

\subsection{Finite-sample considerations}\label{sec:finite}
These results provide intuition for the relative strengths and weaknesses of each approach, which are further demonstrated using real-world data in the following section. Recall that these theoretical results apply to an infinite data setting. In finite data settings where the distribution shift is of larger order than sampling uncertainty, we expect the intuition on the strengths and weaknesses of each estimator to remain consistent. However, if the distribution shift is very small there are cases where $\tau^A$ or $\tilde{\tau}^B$ may outperform $\tau^C$ asymptotically. See Appendix \ref{app:finite} for more details.

\section{EMPIRICAL ASSESSMENTS}

\subsection{Early Childhood Education}

In this application, we use data from the Student Teacher Achievement Ratio project (Project STAR) in Tennessee \citep{SIWH9F_2008}. The dataset contains test scores, demographics, and other information on a large cohort of elementary school students who were tracked by the Tennessee State Department of Education as part of a study focused on evaluating the effect of class size on early student performance \citep{finn1990answers}. 

Our primary outcome, $Y_2$, is students' second-grade scores on the Stanford Achievement Test (SAT), a standardized test with components in math, reading, and listening. We evaluate various proxy outcomes, $Y_1$, including first-grade SAT scores, first-grade and kindergarten scores on the math component, and kindergarten attendance records. Collectively, these proxies range in correlation with the outcome of interest from $0.18$ to $0.81$. 

The covariates $X$ include the students' demographic information and their class size/type, which students were randomized into as part of the study. Our final dataset contains the $3,033$ students for which all our variables are measured. More information on the data and variables can be found in Appendix \ref{app:empirical_educ}.

A relevant policy goal one might have is to predict which students' longer-term outcomes would be most benefited by being placed in smaller class sizes starting in kindergarten. Students currently enrolling in kindergarten, however, may have a distribution shift compared to the previous cohorts whose outcomes we can observe.

To test our proposed method, we randomly split the data into three datasets, corresponding to data from period $0$, $-1$, and $-2$. To induce distribution shift in the relationship between $X$ and $(Y_1,Y_2)$ we randomly permute the covariate vectors for a certain percentage of the period $-1$ dataset. For the period $-2$ dataset, we perform this random permutation twice to induce a stronger shift. We then implement our proposed methods on the resulting data, using random forests to estimate our models, and averaging the results over 1000 random data splits. 

\begin{figure}[ht!]
\begin{center}
\includegraphics[width=1\textwidth]{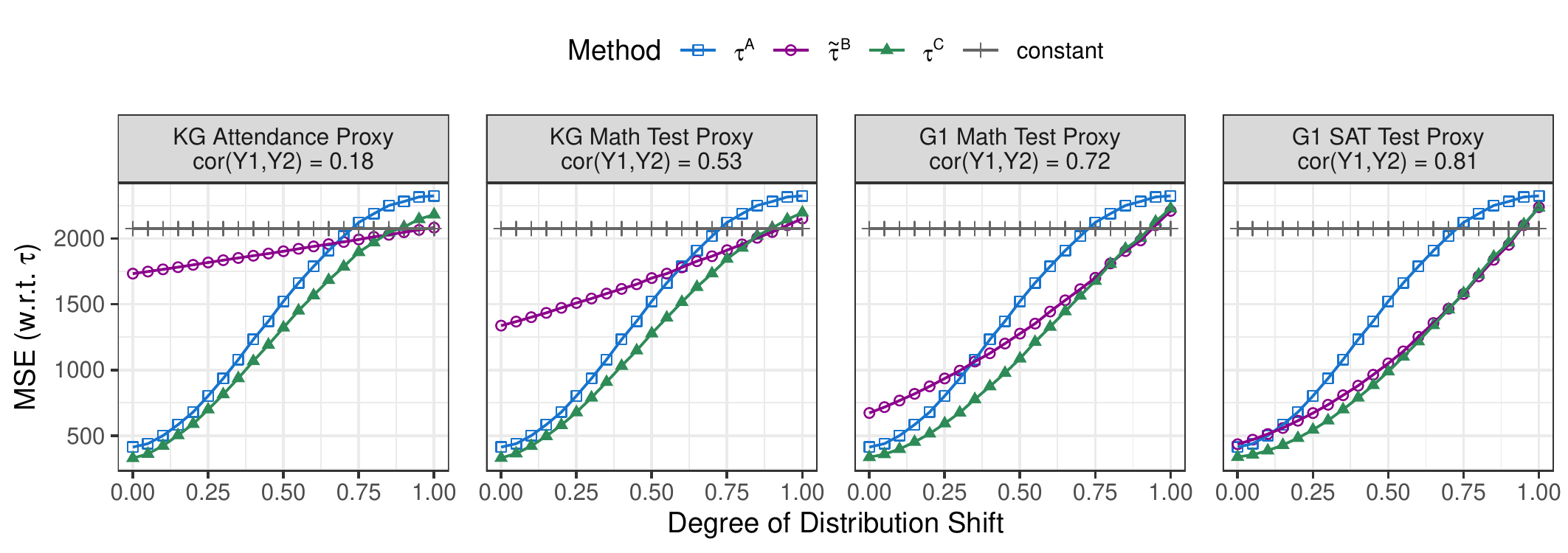}
\caption{MSE of the alternative strategies $\tau_A$, $\tilde{\tau}_B$, and $\tau_C$ in predicting second-grade SAT test scores, with varying degrees of distribution shift and proxy strength (measured by its correlation with $Y_2$). In the panels, KG corresponds to ``Kindergarten'' and G1 corresponds to ``first grade''. For more details see Appendix \ref{app:empirical_educ}.} \label{fig:educ_mse}
\end{center}
\end{figure}

Figure \ref{fig:educ_mse} shows the results, with each panel using a different proxy variable (for $\tilde{\tau}_B$ and $\tau_C$). The $y$-axis corresponds to the MSE as defined in Theorem \ref{thm:mse}, and the $x$-axis corresponds to the degree of distribution shift, from 0 (no permutation) to 1 (100\% permutation). MSE curves are shown for each of the methods, along with that of an intercept/constant model as a reference point. Under all scenarios, $\tau_C$ performs better than or approximately as well as $\tau_A$ and $\tilde{\tau}_B$. The performance of all methods degrades as the strength of the shift increases. As in the theoretical results, $\tau_A$ and $\tau_C$ perform (roughly) equivalently under no shift, but $\tau_C$ degrades more slowly than $\tau_A$ as shift increases. In addition, $\tau_C$  outperforms $\tilde{\tau}_B$ under no shift, and the two methods' performance converges as shift increases. Taken together, these results show the robustness property of $\tau_C$. Also in line with the theory, the advantage of $\tau_C$ over $\tau_A$ ($\tilde{\tau}_B$) grows (shrinks) with a stronger proxy relationship. Finally, it is also worth noting that as the shift gets stronger, it can eventually reach a point where none of the methods can learn anything useful.

Because the outcome $Y_2$ is inherently noisy, it is difficult to discern the practical improvement that $\tau_C$ offers relative to $\tau_A$ or $\tilde{\tau}_B$ based solely on Figure \ref{fig:educ_mse}. For that reason, Appendix \ref{app:empirical_educ} shows the same plot but with R-squared on the $y$-axis, thus demonstrating the superior predictive power of $\tau_C$. For example, using the proxy with a correlation of $0.72$, the predictive performance of $\tau_C$ is more than twice as strong as that of $\tau_A$ when the distribution shift surpasses $0.5$.

\subsection{Asylum Seeker Assignment}

In this section, we demonstrate the performance of the proposed approaches on asylum seeker data from the Netherlands. This evaluation, which uses proprietary, sensitive data, was approved by the Institutional Review Boards at UC Berkeley, Harvard, and Stanford.

The data consists of background characteristics, arrival year/month, assigned location (corresponding to one of the 35 labor market regions within the Netherlands), and employment outcomes for adult asylum seekers in the Netherlands across multiple years. The modeling approach used to estimate $\tau^A$, $\tau^B$, and $\tau^C$ is based on the methodology developed in \citet{bansak2018improving}. In this approach, separate stochastic gradient boosted trees are fit for each labor market. We note that, because of the quasi-random status quo assignment procedure that generated the training data, the predictions in this setting can be interpreted causally. These predictions are then used in an algorithm (such as those developed in \citep{bansak2022outcome, ahani2021dynamic}) designed to suggest an optimal labor market region for each incoming family.

For this assessment, we consider the following two outcomes: $Y_2$ is the proportion of time worked in an asylum seeker's first two years after assignment, and $Y_1$ is the proportion of time worked in the first year after assignment. Based on preliminary analyses of the data, we find strong evidence of distribution shift in the data and a strong proxy relationship between $Y_1$ and $Y_2$ (see Appendix \ref{app:empirical_asylum}).

The asylum seekers who arrived between 2018-06-01 and 2018-08-31 are designated as the period 0 ``test cohort'', and we employ the proposed approaches as if these families were arriving in the present time, with the goal of maximizing average two-year employment. For each individual in the test cohort, and for each labor market region, we generate predictions using $\tau^A$, $\tau^B$, and $\tau^C$. We note that given the test cohort's start date, estimation of $\tau^A$, $\tau^B$, and $\tau^C$ can only utilize data on $Y_2$ prior to 2016-06-01 and $Y_1$ prior to 2017-06-01 (see Appendix \ref{app:empirical_asylum} for more estimation details). 

Because the predictions are used in to inform the downstream assignments, we will evaluate 1) the predictive accuracy of each method, and 2) the expected impact of the varying methods, taking into account the downstream assignment problem. Regarding evaluation (1), as the theory suggests, $\tau_C$ results in a 20-25\% smaller MSE than $\tau_A$ and $\tilde{\tau}_B$ (see Appendix \ref{app:empirical_asylum}). 

\begin{figure}[ht!]
\begin{center}
\includegraphics[width=1\textwidth]{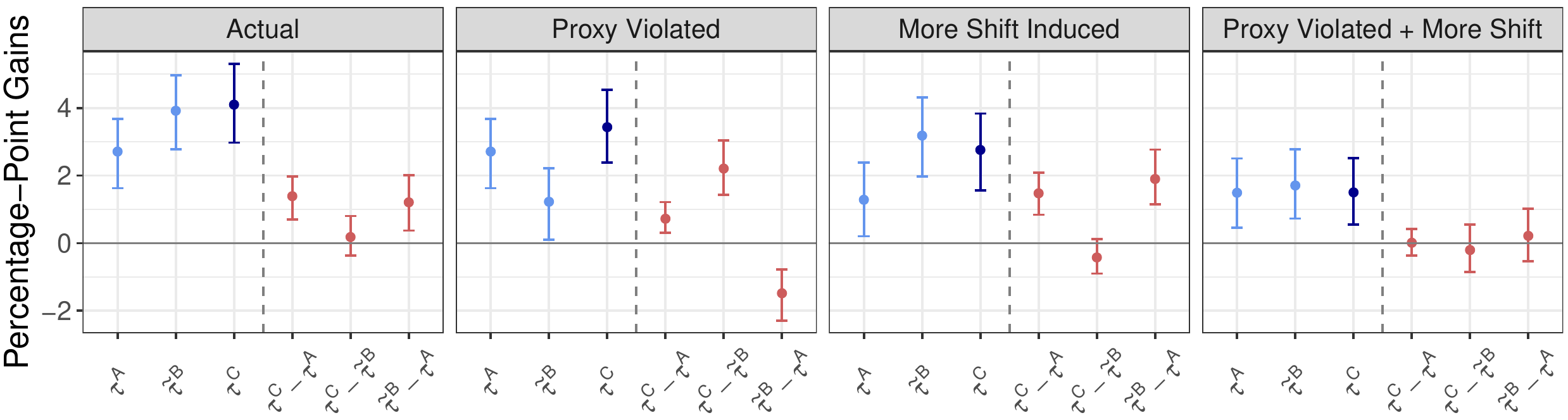}
\caption{Counterfactual gains in two-year employment (compared to status quo employment rate of 14.26\%) with actual data (left) and perturbed data using the alternative strategies $\tau_A$, $\tau_B$, and $\tau_C$. Confidence intervals are 95\% bootstrapped. For details see Appendix \ref{app:empirical_asylum}.} \label{fig:impact_all}
\end{center}
\end{figure}

Evaluation (2) is critical: even if $\tau_C$ attains superior predictive accuracy, if all three methods result in similar geographic assignments then they will also result in similar two-year employment for the test cohort. To that end, Figure \ref{fig:impact_all} (left panel) shows the predicted impact of each method, in terms of gains in two-year employment relative to the status quo, when geographic assignments are made on the basis of the predictions attained by $\tau_A$, $\tilde{\tau}_B$, and $\tau_C$. For more details on the impact evaluation, see Appendix \ref{app:empirical_asylum}.

The gains in two-year employment are largest using the $\tau^C$ estimator, closely followed by $\tilde{\tau}^B$, followed by $\tau^A$. These results suggest that both the distribution shift between period -2 and 0 is relatively large, and one-year employment is a good proxy for two-year employment. Thus, not only does the $\tau^C$ estimator result in the best accuracy, but these improved predictions translate into meaningful gains in employment.

To further illustrate the theory, we perform additional tests in which we induce (a) a violation of the proxy relationship, and (b) additional distribution shift. For the proxy violation test, $Y_1$ is randomly permuted for 50\% of the full data set, thereby weakening the relationship between $Y_1$ and $Y_2$. For the additional distribution shift test, the covariate vectors are randomly permuted for 50\% of the data prior to 2016-06-01. For each test, the permutations are applied prior to the training of the models corresponding to $\tau^A$, $\tau^B$, and $\tau^C$.

The impact on gains in two-year employment for each test is shown in Figure \ref{fig:impact_all}. When the proxy relationship is violated, the performance of $\tau^B$ dramatically suffers, as does the performance of $\tau^C$ to a lesser extent. However, $\tau^C$ still outperforms $\tau^A$, thus demonstrating its robustness to proxy violations. When more distribution shift is present in the data, the performance of all three predictors suffers, and $\tau^B$ and $\tau^C$ perform similarly well (with a slight edge to $\tau^B$, although their difference is not statistically significant). When both a proxy violation and more distribution are introduced, all three estimators perform similarly.

\section{CONCLUSIONS, LIMITATIONS, AND BROADER IMPACT}

In this research, we characterized the strengths and weaknesses---theoretically and empirically---of three approaches for generating predictions in settings with random distribution shift. The standard approach trains a model to directly predict a (long-term) outcome of interest, suffering under distribution shift. The proxy approach directly predicts a short-term proxy outcome, suffering if the proxy relationship is weak. The hybrid approach enjoys the strengths of both the standard and proxy approaches without suffering from their respective failure points. That is, the hybrid approach is robust to both distribution shift and the strength of the proxy relationship.

Our theoretical conclusions were empirical validated in two real-world contexts: asylum seeker resettlement and early childhood education. In these domains, the hybrid approach consistently outperformed both the standard and proxy approaches, demonstrating the practical relevance of our theoretical insights. More broadly, we expect the hybrid approach to perform well in other real-world domains where distribution shifts arise due to natural economic and social changes.

The model considered in this paper has several limitations. We consider a specific type of distribution shift, characterized in Section \ref{sec:theory}. The results of our empirical analyses show that this can be a reasonable model in certain contexts. However, it does prohibit certain dynamics; for example, we do not address situations where $P_{-2} \approx P_0$, but $P_{-1}\neq P_0$. Such a phenomenon could occur if an isolated event only impacts period $-1$. Furthermore, this paper only considers random shifts. There could be large, unpredictable shifts that make all historical data useless (for example a new technology that disrupts an economy).

\begin{ack}
The authors are grateful for the guidance and data access provided by the Central Agency for the Reception of Asylum Seekers (COA) in the Netherlands and Statistics Netherlands (CBS). This work was supported by Stanford University's Human-Centered Artificial Intelligence (HAI) Hoffman-Yee Grant and J-PAL's European Social Inclusion Initiative (ESII).
\end{ack}

\bibliographystyle{plainnat}
\bibliography{references}
\clearpage
\appendix

\section{APPENDIX FOR EMPIRICAL ASSESSMENTS}\label{app:empirical}

\subsection{Early Childhood Education}\label{app:empirical_educ}

\subsubsection{Background}

Experts and policymakers have long been interested in forecasting and understanding the determinants of positive outcomes in early childhood education. Being able to effectively estimate expected outcomes in early childhood education and identify effective interventions can facilitate educational or curricular reforms that improve immediate student learning, while also improving downstream educational and economic outcomes later in life \citep[e.g. see][]{krueger2001effect,duncan2007school,burchinal2008predicting}. However, the relationships among learning outcomes and other variables can shift over time owing to changes occurring in schools, school systems/districts, and the populations and economies linked to those systems/districts.

One prominent example of a study focused on evaluating an early education intervention is the Student Teacher Achievement Ratio project (Project STAR). Conducted in Tennessee in 1985–1989 by the State Department of Education, Project STAR was a longitudinal study focused on evaluated the effects of class size on student academic performance, as measured by a number of standardized and curriculum-based tests \citep{finn1990answers}. We use the data from Project STAR for this application.

\subsubsection{Data and Estimation Procedures}

The Project STAR data set, which is available at \cite{SIWH9F_2008}, contains the test scores, demographics, and other information on a large cohort of elementary school students who participated in the study. The full data set contains information for $11,601$ students, though we focus on the subset of students who have complete observations (i.e. no missing data) for the variables we employ in our analysis ($n = 3,033$). These variables, and their role in our analysis, are as follows:
\begin{itemize}
    \item Primary outcome ($Y_2$)
        \begin{itemize}
            \item[-] Students' second-grade scores on the Stanford Achievement Test (SAT), sum of the math, reading, and listening components
        \end{itemize}
    \item Proxy outcomes ($Y_1$)
        \begin{itemize}
            \item[-] Students' first-grade scores on the SAT, sum of the math, reading, and listening components
            \item[-] Students' first-grade scores on the math component of the SAT
            \item[-] Students' kindergarten scores on the math component of the SAT
            \item[-] Students' kindergarten attendance record
        \end{itemize}
    \item Covariates ($X$)
        \begin{itemize}
            \item[-] Year of birth
            \item[-] Month of birth
            \item[-] Race (White, Black, Asian, Hispanic, Native American, Other)
            \item[-] Gender
            \item[-] Indicator for eligibility for free lunch (a measure of family socioeconomic status)
            \item[-] Indicator for special education status
            \item[-] Randomly assigned class size/type (Regular class, Regular class + teacher's aide, Small class)
        \end{itemize}
\end{itemize}

All models were trained using the \texttt{ranger} package in \texttt{R} using the default parameter settings. 

\subsubsection{Additional Results}

\begin{figure*}[ht!]
\begin{center}
\includegraphics[width=1\textwidth]{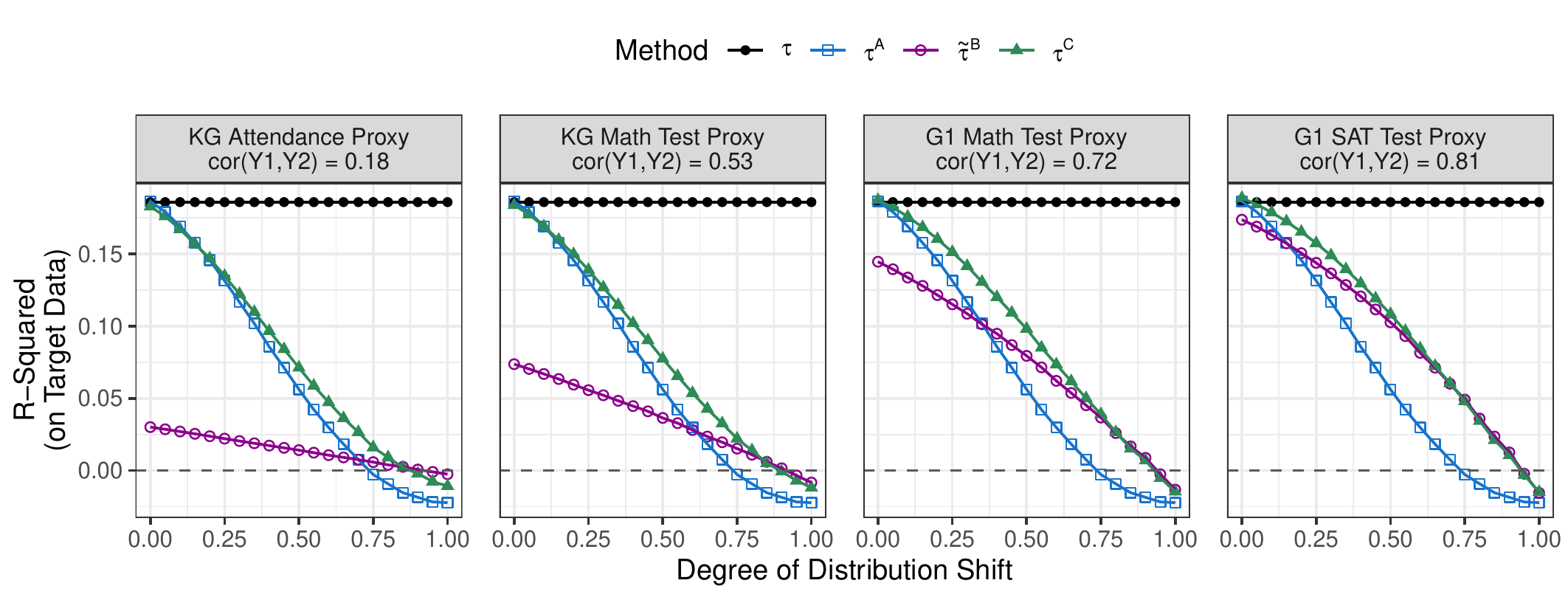}
\caption{R-squared of the alternative strategies $\tau_A$, $\tilde{\tau}_B$, and $\tau_C$ in predicting second-grade SAT test scores, with varying degrees of distribution shift and proxy strength (measured by its correlation with $Y_2$). The R-squared is calculated with respect to the target (i.e. period $0$) data. In the panels, KG corresponds to ``Kindergarten'' and G1 corresponds to ``first grade''.} \label{fig:educ_r2}
\end{center}
\end{figure*}

The outcome $Y_2$ is inherently noisy, and thus the practical improvement that $\tau_C$ offers relative to $\tau_A$ or $\tilde{\tau}_B$ can be challenging to interpret based solely on the MSE results shown in Figure \ref{fig:educ_mse} in the main text. Here, Figure~\ref{fig:educ_r2} shows analogous results but instead using the R-squared, which is more straightforward to interpret from a practical perspective. Based on this figure, we can see clearly the superior predictive power of $\tau_C$. For example, using the proxy with a correlation of $0.72$, the predictive performance of $\tau_C$ is more than twice as strong as that of $\tau_A$ when the distribution shift surpasses $0.5$.

\subsection{Asylum Seeker Assignment}\label{app:empirical_asylum}

\subsubsection{Data}

We use data on asylum seekers in the Netherlands from two data sources. The first is the administrative data on individuals granted temporary asylum residence permit (``permit holders") from the Central Agency for the Reception of Asylum Seekers (COA), which is the government agency in charge of the asylum system in the Netherlands. This is a comprehensive data set covering all permit holders in the Netherlands containing their background characteristics, procedural information, and location data. The second is the Asielcohort microdata compiled by Statistics Netherlands (CBS). The Asielcohort microdata is comprised of merged data from a number of administrative databases, several of which track various measures of permit holders in the Netherlands after arrival. These include economic, educational, and other indicators of integration and well-being that can be evaluated as downstream outcomes.

The target population for our assessment is comprised of permit holders who were geographically assigned in the Netherlands through the regular housing procedure. Further, we only consider data and outcomes for adults (i.e. 18 years or older). We also exclude some subsets of permit holders who fall outside the scope of the objectives of algorithmic assignment or for whom data are unreliable.\footnote{We exclude permit holders who fell under the 2019 Children’s Pardon, resettlers / relocants / asylum seekers covered by the EU-Turkey deal, and permit holders who have ``hard criteria'' that pre-determined their locations. Beyond family reunification constraints, hard criteria may be motivated by other determinants such as employment contracts and medical issues.} In addition, the target population for algorithmic assignment further excludes family reunifiers, since such permit holders are automatically reunified with their family in the Netherlands.\footnote{Note, however, that data for family reunifiers is included the data used to estimate the various models comprising $\tau^A$ and $\tau^B$, with an family reunifier indicator also included as a variable in the models. Family reunifiers were not, however, included in the ``test cohort" defined below since they are outside the scope of algorithmic assignment.} The permit holders in the available data were assigned between January 2014 and August 2018 ($n \approx 46,000$).

As per rules on usage of these data sets, and in accordance with our data access and use agreements with CBS, our access to and analysis of these data is conducted entirely via a secure Remote Access Environment. All statistics and results that are reported from these data are checked and cleared for export by CBS.

\subsubsection{Set-up}
For the refugee and asylum seeker assignment problem, the formal problem set-up is as follows. We note that it closely follows the set-up in the main text, with the key distinctions being 1) location-specific outcomes for every unit, and 2) a formal presentation of the causal model for this application. 

In each period, refugees arrive and must be assigned to a single location within the set $\mathcal{L} = \{1, 2, ... |\mathcal{L}|\}$. Let $J \in \mathcal{L}$ denote the location assignment of a refugee that arrives in the current period, and again let $X$ denote a vector of background characteristics. 

Following the potential outcomes framework of \cite{neyman1923applications} and \cite{rubin1974estimating}, let $Y_1(j)$ and $Y_2(j)$ denote the one-period and two-period potential outcomes under assignment to location $j \in \mathcal{L}$. We assume that the stable unit treatment assumption (SUTVA) holds, so
we observe $Y_1 = Y_1(J)$ and $Y_2 = Y_2(J)$ \citep{rubin1980randomization}. 

For refugees who arrive in any period $t$ and the variables defined above, we posit the existence of a tuple generated according to a probability distribution $P_t$:
\begin{equation*}
(Y_1(1),...,Y_1(|\mathcal{L}|),Y_2(1),...,Y_2(|\mathcal{L}|),X,J)_{t} \sim P_t.
\end{equation*}
Further, we assume that within any period, the assignment mechanism is such that
\begin{equation*}
Y_1(j) \indep J \: | \: X, \;\; Y_2(j) \indep J \: | \: X, \;\; \text{and}\;\; P(J = j | X = x) > 0 \:\:\: \forall \:\:\: j, x.
\end{equation*}
This is often referred to as the ignorability assumption \citep{rosenbaum1983central}.
It is satisfied here when location assignments are made on the basis of the observed characteristics, and when those characteristics do not preclude refugees from being assigned to any particular location, as in our application to the Dutch asylum system. Therefore,
$\mathbb{E}_t[Y_k(j) | X] = \mathbb{E}_t[Y_k | X, J=j]$ for $k=1,2$. This paper is focused on the identifiable quantities $\mathbb{E}_t[Y_k | X, J=j]$.

To maximize $Y_2$, as in the main text, we would ideally choose location assignments for the refugees arriving in period $0$ by first estimating
$\tau(x,j) = \mathbb{E}_0[Y_2 | X=x, J=j]$, and using these estimates within a static or dynamic matching procedure to optimally assign each refugee to a particular location subject to any necessary constraints, such as capacity constraints and the need to keep family members together, as in the algorithmic assignment procedures proposed by \cite{bansak2018improving, ahani2021placement, ahani2021dynamic} and \cite{bansak2022outcome}. 

However, we do not observe $Y_2$ (nor do we observe $Y_1$) for the cohort that arrives in period $0$ at the time of their assignment. Instead, at the time of assignment, we observe the data $(Y_2, Y_1, X, J)_{-2} \sim P_{-2}$, $(Y_1, X, J)_{-1} \sim P_{-1}$, and $(X)_{0} \sim P_{0}$. As described in the main text, we can thus employ the three proposed strategies $\tau_A$, $\tau_B$ and $\tau_C$ as described. The only subtle difference is that, for every unit, each strategy results in $\mathcal{L}$ predictions---one for each possible assignment location.

\subsubsection{Modeling and Estimation Procedures}

The modeling approach is based on the methodology developed in \citet{bansak2018improving}. We first merge the historical data for the target population's background characteristics, employment outcomes, and geographic locations. Using supervised learning on these merged data, we fit separate models across each labor market region (LMR) that predict one- or two-year employment using the predictors described in a section below. To do so, for each model we first subset the training data to permit holders who were assigned to a given LMR, and then use this subset for the model training. The next subsection describes the basis upon which LMRs were targeted as the geographic locations of interest.

To generate our models, we employ stochastic gradient boosted trees with squared error loss, which we implement in \texttt{R} using the \texttt{gbm} package. We employ cross-validation to determine the optimal values for tuning parameters. Specifically, we cross-validate over the number of boosting iterations (trees) and the interaction depth of the trees. We employ 5 folds in our cross-validation, allowing the \texttt{gbm} functionality to determine the random splits. For each model, we cross-validate over tree depths of 3-8 and a number of trees that we ensure are sufficient to be able to identify the number that yields the minimum CV mean squared error (i.e. we ensure that we do not consider too few trees to find the minimum CV error). Parameters were tuned independently for each location-specific model. Preliminary assessments on the bag fraction, learning/shrinkage rate, and minimum number of observations per node demonstrated relatively little to no impact on model performance within conventional value ranges, and so these parameters were held fixed at 0.5, 0.01, and 5, respectively. Where appropriate, 95\% confidence intervals are generated using a nonparametric bootstrap.

\subsubsection{Target Geography}

Selecting the appropriate target unit of geography for algorithmic recommendations requires balancing three goals: (1) generating a large number of geographic options to provide the algorithmic recommendation procedure with as much geographic variation as possible, (2) ensuring that each geographic option is associated with a sufficient amount of historical data such that accurate and effective predictive models can be trained, and (3) identifying levels or regions of geography that are administratively compatible with the underlying goals and procedures. 

Based on these criteria, and in consultation with our partners at COA, we determined that the best target unit of geography in the Dutch context is the labor market region (LMR). Hence, as noted above, we generate separate predictive models for each of the 35 LMRs, and the algorithmic assignment procedure determines the optimal LMR for each family of permit holders (i.e. the LMR where they have the highest chance of employment success, subject to all the constraints). 

\subsubsection{Predictors}

Based on data availability, the pre-arrival characteristics we include as the covariates $X$ are age, gender, martial status, prior education, number of family members, country of origin, religion, native language, ethnicity, and prior work experience and industry. In addition, we also include in $X$ several key variables related to the assignment procedure as predictors, including the month of assignment, year of assignment, whether or not someone is a family reunifier, and what type of processing location their housing interview took place at. 

\subsubsection{Evidence of proxy relationship and distribution shift} \label{app:proxy_shift_evidence}

As a simple analysis of the proxy relationship, we fit a linear regression of $Y_2$ on $Y_1$ using the entire dataset and find a large $R^2$ value of $0.625$. The $R^2$ increases to $0.642$ when the covariates are added to the model. 

To establish the presence of distribution shift, we perform two analyses making use of the data from 2015-06-01 to 2016-05-31 (analogous to period $-2$ in our impact assessment) and from 2016-06-01 to 2017-05-31 (analogous to period $-1$). First, we fit classifiers (using gradient boosted trees according to the procedures described above, but with binomial deviance loss in place of squared error loss) that predict which period each data point belongs to as a function of the outcomes and the covariates (described above). Note that for this procedure, we omit the month and year covariates for obvious reasons. We find solid predictive performance with these models: a classification accuracy of $0.768$ (relative to $0.547$ under a null/intercept model) and an $R^2$ of $0.344$. 

Second, we split the period $-1$ data into training and test sets, and we estimate models (again using gradient boosted trees according to the procedures described above, and with squared error loss) that predict the two-year outcome as a function of the covariates. We fit models using the period $-1$ training data, and separately also fit models using a period $-2$ training set that is randomly sampled to have identical size as the period $-1$ training set. We then compare the performance of these models in predicting onto the period $-1$ test set, and we find clearly superior performance of the period $- 1$ training models (a test set $R^2$ of $0.126$) over the period $- 2$ training models (a test set $R^2$ of $0.081$). We find similar results when we use these models to predict onto data that is analogous to period $0$.

\subsubsection{Application}

Let $\bm{W}^{A}$, $\bm{W}^{B}$, and $\bm{W}^{C}$ be the prediction matrices corresponding to $\tau_A$, $\tau_B$, and $\tau_C$, where each row corresponds to an individual in the test cohort, and columns corresponding to labor market regions.

$\bm{W}^{A}$ is estimated by fitting models for expected two-year employment using the data prior to 2016-06-01 (i.e. two years earlier than the test cohort), equivalent to period $-2$, and then applying those models to the test cohort. $\bm{W}^{B}$ is estimated by fitting models for expected one-year employment using the data prior to 2017-06-01, equivalent to the union of period $-1$ and $-2$, and then applying those models to the test cohort. $\bm{W}^{C}$ is estimated via a two-step process. In the first step, models for expected two-year employment are fit using using the data from prior to 2016-06-01, with one-year employment being included as a regressor. Those models are applied to the data between 2016-06-01 and 2017-05-31. With these predicted/expected values in hand, in the second step, another set of models for two-year employment are fit using all data prior to 2017-06-01---which can be done ``feasibly" by using the predicted/expected values for two-year employment as the outcome for these models---with only the variables contained in $X$ as the regressors. In addition, the data from prior to 2016-06-01 are also included in the training of these models, with the true/observed values (rather than predicted/expected values) of two-year employment used for those observations. This methodology effectively combines period $-1$ and $-2$ for the $\tau^B$ estimator and for the second stage of the $\tau^C$ estimator, which is a deviation from the assumptions made in Section \ref{sec:theory}. One could also estimate $\tau^B$ and the second stage of $\tau^C$ using only period $-1$ data (between 2016-06-01 and 2017-05-31). In Section \ref{app:results-odd} we show a figure analogous to Figure \ref{fig:impact-odd} where $\tau^B$ and $\tau^C$ were estimated using this approach.

Let $\bm{W}^{*}$ denote the ground truth prediction matrix, estimated by fitting models for expected two-year employment using \emph{all} of the available data, including the data for the 2018-06-01 -- 2018-08-31 test cohort. We use all of the data for this quantity for the sake of counterfactual evaluation; if one were trying to assign the 2018-06-01 -- 2018-08-31 in the real world (i.e. if that was the present), it would obviously not be possible to estimate $\bm{W}^{*}$. In contrast, $\bm{W}^{A}$, $\bm{W}^{B}$, and $\bm{W}^{C}$ are all estimated in ways that would be possible, allowing them to be used to determine the counterfactual assignment decisions. 

\subsubsection{Assignment and Constraints}
Once the models corresponding to $\tau^A$ and $\tau^B$ have been estimated, the models can then be applied to the test cohort to estimate their expected employment success (based on $Y_2$ or $Y_1$, respectively for $\tau^A$ and $\tau^B$) at each of the possible LMRs. Optimal algorithmic assignment decisions on which specific LMR each permit holder should be assigned to can then be made, based on maximizing the expected average employment subject to the constraints. There are two key types of assignment constraints that we take into account for our assessment.
The first are family-related constraints. All permit holders in the same family (or associated with the same case number for other reasons) must be assigned together to the same location. The second are location capacity constraints. That is, permit holders must be assigned across labor market regions according to pre-determined capacity and proportionality guidance. 

Let $\bm{Z^A}$, $\bm{Z^B}$, and $\bm{Z^C}$ be defined as the (predicted) optimal assignments corresponding to each method, given by:
\begin{align*}
	\bm{Z^k} = \arg\max_{\mathbf{z}\in \mathcal{Z}} \sum_{i=1}^{n_0}\sum_{j=1}^{|\mathcal{L}|} w^{k}_{ij} z_{ij}^k \; \text{ for $k=A,B,C$}
\end{align*}
where $\mathcal{Z}$ is the set of feasible assignments of asylum seekers to labor market regions. At a minimum, $z_{ij}$ must be binary with each asylum seeker assigned to exactly one location, and $\mathbf{Z}$ must satisfy the constraints.

To compare the impact of each method, we will compare the quantities: 
\begin{align*}
	E^k &= \sum_{i=1}^{n_0}\sum_{j=1}^{|\mathcal{L}|} w^{*}_{ij} z^k_{ij} \text{ for $k=A,B,C$}
\end{align*}

\subsubsection{Results}
 The mean squared error for method $K$ is computed by taking the average over locations $j \in \mathcal{L}$ and refugee strata $x$. The MSE for each test is shown in Table 1.

\begin{table}[ht!] \label{tab:placebo-MSE}
\begin{center}
\begin{tabular}{l  c c c}
     \textbf{SETTING}   & $\mathbf{\tau^A}$ & $\mathbf{\tilde{\tau}^B}$ & $\mathbf{\tau^C}$ \\
        \midrule
 Actual & 0.010084 & 0.009388 & 0.007520 \\  
 Proxy Violated & 0.010084 & 0.013649 & 0.009756 \\   
 More Shift & 0.012559 & 0.009721 & 0.008650 \\  
 Proxy Violated + More Shift & 0.012407 & 0.013527 & 0.012243 \\   \\
\end{tabular}
\caption{MSE Results}
\end{center} 
\end{table}

\subsubsection{Results where old data is dropped for $\tau^B$ and $\tau^C$}\label{app:results-odd}
\begin{figure}[ht!]
    \centering
    \includegraphics[width=\textwidth]{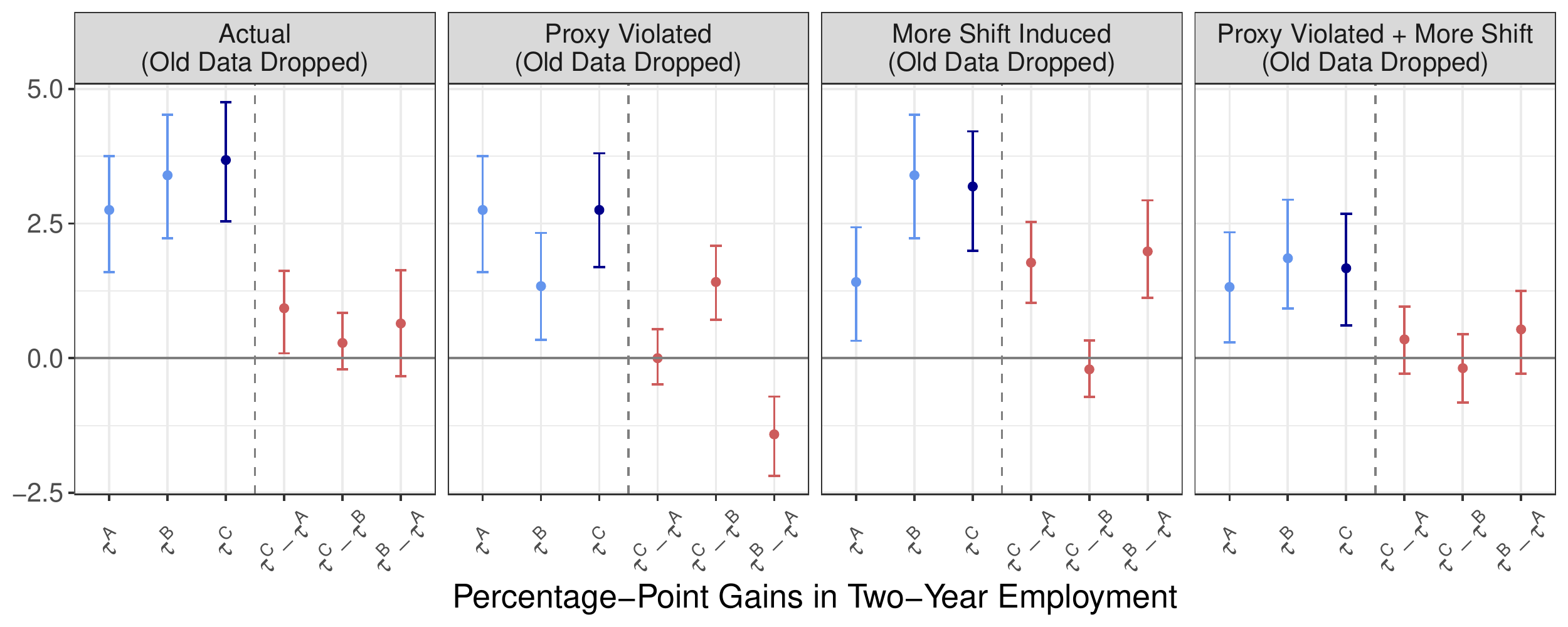}
    \caption{Counterfactual gains in two-year employment (compared to status quo) with actual data (left) and perturbed data using the alternative strategies $\tau_A$, $\tau_B$, and $\tau_C$. Confidence intervals are 95\% bootstrapped.}
    \label{fig:impact-odd}
\end{figure}

Figure \ref{fig:impact-odd} is analogous to Figure \ref{fig:impact_all} but shows the results where $\tau^B$ and the second stage of $\tau^C$ are estimated only using data from between 2016-06-01 and 2017-05-31. This method more closely aligns with the theoretical results, but practically turns out to be less desirable because the sample size becomes much smaller. As can be seen by comparing Figure \ref{fig:impact-odd} and \ref{fig:impact_all}, both $\tau^B$ and $\tau^C$ perform slightly worse using this methodology. \\

\section{EXTENSIONS}\label{app:extensions}

\subsection{More than two outcomes}

In a more general version of the problem discussed above, we have $T+1$ time periods: $t\in \{0,-1,...,-T\}$, where period $0$ is the present time, and we have $T$ outcomes, $Y_1$ through $Y_T$, with $Y_T$ being the policy outcome of interest. The period $t$ data contains outcomes $Y_1$ through $Y_t$.  There are now \emph{at least} $T+T(T+1)/2$ strategies one could pursue in choosing an outcome and estimation method to guide the algorithmic recommendations. In particular, there are $T$ estimands of the form: $\tau^t(x)=\mathbb{E}_{-t}[Y_t|X=x]$ for $t=1,...,T$. There are also $T(T+1)/2$ nested methods of the form: $$\tau^{t_1-t_2}(x)=\mathbb{E}_{-t_1}[\mathbb{E}_{-t_1-1}[...\mathbb{E}_{-t_2+1}[\mathbb{E}_{-t_2}[Y_{t_2}|Y_{t_1},...,Y_{t_2-1},x]|Y_{t_1},...,Y_{t_2-2},x]...]|x]$$ for $t_2>t_1$.
For example, when $T=3$, 
$$\tau^{1-3}(x) = \mathbb{E}_{-1}[\mathbb{E}_{-2}[\mathbb{E}_{-3}[Y_3|Y_2,Y_1,x]|Y_1,x]|x].$$ Similar intuition for the relative merits and drawbacks of each method as in Section \ref{sec:theory} will hold. In particular, we expect $\tau^{1-T}(x)$ to have similar robustness properties as $\tau^C$.

\subsection{Other data fusion problems} \label{app:other_fusion}
In addition to extending the modeling framework to an arbitrary number of time periods, one could also extend the framework to non-temporal shifts. In particular, consider a scenario with three datasets, I, II, and III. Our goal is to predict a particular outcome, outcome A, for individuals in dataset I, where no outcome labels are present. Dataset II consists of a population of individuals and an outcome B, related to outcome A. Finally, dataset III contains yet another population for which outcomes A and B are both present. The question is how to use datasets II and III, along with the outcomes they contain, to best predict outcome A for dataset I. In this paper, the random distribution shift model effectively meant that period -1 data is closer to period 0 than period -2 is to period 0. In this generalization, a key difficulty is measuring the ``closeness'' of datasets I, II, and III. If dataset II is closer to dataset I than dataset III is to I, one could employ an analogous estimator to estimator $\tau^C(x)$.

\section{FINITE-SAMPLE CONSIDERATIONS} \label{app:finite}

In the following we will study how the procedures compare from a asymptotic viewpoint, in the absence of distribution shift (that is, if $P_{-2} = P_{-1} = P_0$). This setting is particularly important for relatively small shifts, since in this case the error due to sample size will dominate the overall mean-squared error. For mathematical simplicity, we assume that the ratio of units for the two periods converges to a constant, that is $\frac{n_{-1}}{n_{-2}} \rightarrow \rho \in (0,\infty)$. We write $n = n_{-1} + n_{-2}$.

In general, it is difficult to compare the three approaches since the performance depends both on the choice of machine learning algorithms and on the data set. In particular, similar to "no free lunch" theorems  in machine learning \citep{wolpert1996lack}, we expect that for any choice of algorithm, any of the strategies can perform best depending on the dataset. That being said, in the following, we will see that in the setting where $X$ and $Y_1$ is discrete, a clear comparison can be drawn. In this case, in low-dimensional settings the estimator can be based on sample proportions. Let $I_{-2,x}$ be the set of indices for units from period $t=-2$ that have $X_i = x$. 
\begin{align*}
    \hat \tau^A(x) = \frac{1}{ \# I_{-2,x}} \sum_{i \in I_{-2,x}} Y_{2,i}.
\end{align*}
We now define an estimate of $\tau^C$. To this end, let $I_{-1,x}$ denote the set of indices for units from time period $t=-1$ that have $X_i=x$. Similarly, let  $I_{-2,x,y}$ be the set of indices for units from time period $t=-2$ that have $X_i = x$, and $Y_{1,i} = y$.
As above, we can define
\begin{align*}
\hat \tau^C(x) &= \frac{1}{ \# I_{-1,x}} \sum_{i \in I_{-1,x}} \hat Q(x,Y_{1,i}), \, \, \text{ where } \, \, \hat Q(x,y) = \frac{1}{ \# I_{-2,x,y}} \sum_{i \in I_{-2,x,y}} Y_{2,i}.
\end{align*}
In the following, we introduce an estimator for $\tau^B$. Analogously to the above, define
\begin{align*}
    \hat \tau^B = \frac{1}{ \# I_{-1,x}} \sum_{i \in I_{-1,x}} Y_{1,i}.
\end{align*}
Let us now compare the asymptotic performance of the three approaches. The estimator $\hat \tau^B$ is justified if the relationship between the target $Y_2$ and the proxy outcome is linear, that is if $\mathbb{E}_{-2}[Y_2|X=x] = \beta_1 \mathbb{E}_{-2} [Y_1 | X=x] + \beta_0$. This is a very strong assumption. In particular, it assumes that the relationship does not change for different values of $X=x$. The proof of the following result can be found in Section~\ref{sec:proofs}.
\begin{example}[Invariant distribution; proxy assumption does not hold]\label{ex:noproxynoshift} Assume that there is no shift, i.e.\ that $P_{-2}(\cdot|x) = P_{-1}(\cdot|x) = P_{0}(\cdot|x)$.
Then, 
\begin{align*}
    &\sqrt{n}( \hat \tau^A(x) - \tau(x) ) \rightharpoonup \mathcal{N}(0, \sigma_A^2) \\
    &\sqrt{n}( \hat \tau^C(x) - \tau(x) ) \rightharpoonup \mathcal{N}(0, \sigma_C^2),
\end{align*}
where $\sigma_A^2 < \sigma_C^2$ if and only if $\rho < 1$. Furthermore, if the linear proxy assumption does not hold, we have
 \begin{equation*}
     \beta_1 \hat \tau^B(x) + \beta_0 - \tau(x) \not\rightarrow_P 0.
 \end{equation*}
\end{example}

Please note that we expect $n_{-2} > n_{-1}$, since for $P_{-2}$ we can pool many observations from previous timepoints. Thus, if there is not distribution shift and the proxy assumption is violated, one would generally prefer $\hat \tau^A$.
If the linear proxy assumption holds and $n_{-1} > n_{-2}$, then the conclusions change. This will be discussed in the following.

The following example shows that in this case $\tau^B$ is preferable over the other procedures. The proof can be found in Section~\ref{sec:proofs}.
\begin{example}[Invariant distribution; proxy assumption holds]\label{ex:proxy} Assume that there is no shift, that is $P_{-2}(\cdot|x) = P_{-1}(\cdot|x) = P_0(\cdot|x)$ and that $\rho > 1$. If $\mathbb{E}_{-2}[Y_2|Y_1,X] = \beta_1 Y_1 + \beta_0$, then for $\hat \tau_{\text{scaled}}^B(x) = \beta_1 \hat \tau^B + \beta_0$ we get
\begin{align*}
        &\sqrt{n}( \hat \tau^A(x) - \tau(x) ) \rightharpoonup \mathcal{N}(0, \sigma_A^2) \\
    &\sqrt{n}( \hat \tau^C(x) - \tau(x) ) \rightharpoonup \mathcal{N}(0, \sigma_C^2) \\
    &\sqrt{n}( \hat \tau_{\text{scaled}}^B(x) - \tau(x) ) \rightharpoonup \mathcal{N}(0, \sigma_B^2).
\end{align*}
Furthermore,
\begin{equation*}
    \sigma_B^2 < \min(\sigma_A^2, \sigma_C^2).
\end{equation*}
\end{example}

\section{PROOFS}\label{sec:proofs}

\subsection{Proof of Theorem~\ref{thm:mse}}

\begin{proof}
For notational simplicity, without loss of generality, we will assume that $Y_1$ is already transformed, that is $Y_1 = \tilde Y_1$.
First, let us prove a few auxiliary results.
By Assumption~\ref{ass:centered}, for fixed $y_2,y_1,y_2',y_1',x,x'$ we have
\begin{equation}\label{eq:first}
   \mathbb{E}[ S_{t}(y_2,y_1|x) ] = 0 \text{ and } \text{Cov}( S_{-2}(y_2,y_1|x), S_{-1}(y_2',y_1'|x')) = 0,
\end{equation}
for $t \in \{-2,-1\}$. For every function $f$ and fixed $x$ we have
\begin{align}\label{eq:first-step}
\begin{split}
    &\text{Var}( \sum_{y_1,y_2} f(y_1,y_2,x) S_{-2}(y_1,y_2|x)  ) \\
    &= \sum_{y_1,y_2} \sum_{y_1',y_2'} f(y_1,y_2,x) f(y_1',y_2',x) \text{Cov}(S_{-2}(y_1,y_2|x),S_{-2}(y_1',y_2'|x))
\end{split}
\end{align}
Here, the variance is over the randomness in the shift $S_{-2}$. Now if $y_1=y_1'$ and $y_2 = y_2'$, then by Assumption~\ref{ass:scaling},
\begin{align*}
    &\text{Cov}(S_{-2}(y_1,y_2,|x),S_{-2}(y_1',y_2'|x)) \\
    &= \text{Var}(S_{-2}(y_1,y_2|x) ) \\
    &= \kappa_{-2}   P_{-2}(y_1,y_2|x) (1 - P_{-2}(y_1,y_2|x)).
\end{align*}

If $y_1 \neq y_1'$ or $y_2 \neq y_2'$, let $A$ be the event that $(y_1',y_2') = (Y_1,Y_2)$ or $(y_1,y_2) = (Y_1,Y_2)$. Then we can use additivity, that is $P_{-1}(A  |X=x  ) = P_{-1}( (y_1',y_2') = (Y_1,Y_2)|X=x) + P_{-1}( (y_1,y_2) = (Y_1,Y_2) | X=x)$ to obtain 
\begin{align*}
    &\text{Var}( P_{-1}(A  |X=x  ))  \\
    &= \text{Var}( S_{-2}(y_1',y_2'|x)) + 2 \text{Cov}(S_{-2}(y_1,y_2|x),S_{-2}(y_1',y_2'|x))  + \text{Var}( S_{-2}(y_1,y_2|x)   )
\end{align*}
Applying Assumption~\ref{ass:scaling} directly on both sides, we get
\begin{align*}
    &\kappa_{-2} P_{-2}(A|x) ( 1 -P_{-2}(A|x))  \\
    &= \kappa_{-2} P_{-2}(y_1,y_2|x) ( 1- P_{-2}(y_1,y_2|x))  + 2 \text{Cov}(S_{-2}(y_1,y_2|x),S_{-2}(y_1',y_2'|x)) \\
    &+ \kappa_{-2} P_{-2}(y_1',y_2'|x) ( 1- P_{-2}(y_1',y_2'|x)).
\end{align*}
Applying additivity to the left-hand side, we get
\begin{align*}
    &\kappa_{-2} ( P_{-2}(y_1,y_2|x) + P_{-2}(y_1',y_2'|x) ) ( 1 -P_{-2}(y_1,y_2|x) - P_{-2}(y_1',y_2'|x) )  \\
    &= \kappa_{-2} P_{-2}(y_1,y_2|x) ( 1- P_{-2}(y_1,y_2|x))  + 2 \text{Cov}(S_{-2}(y_1,y_2|x),S_{-2}(y_1',y_2'|x)) \\
    &+ \kappa_{-2} P_{-2}(y_1',y_2'|x) ( 1- P_{-2}(y_1',y_2'|x)).
\end{align*}
Simplifying, we get
\begin{equation*}
     - 2 \kappa_{-2}  P_{-2}(y_1',y_2'|x) P_{-2}(y_1,y_2|x) = 2 \text{Cov}(S_{-2}(y_1,y_2|x),S_{-2}(y_1',y_2'|x)).
\end{equation*}
Thus,
\begin{align*}
    \text{Cov}(S_{-2}(y_1,y_2|x),S_{-2}(y_1',y_2'|x)) =  - \kappa_{-2} P_{-2}(y_1,y_2|x) P_{-2}(y_1',y_2'|x).
\end{align*}
To summarize, we have
\begin{align*}
    &\text{Cov}(S_{-2}(y_1,y_2,|x),S_{-2}(y_1',y_2'|x)) \\
    &= \begin{cases}
     \kappa_{-2}   P_{-2}(y_1,y_2|x) (1 - P_{-2}(y_1,y_2|x)) &\text{ if } (y_1,y_2) = (y_1',y_2'), \\
      - \kappa_{-2}  P_{-2}(y_1,y_2|x) P_{-2}(y_1',y_2'|x) &\text{ if } (y_1,y_2) \neq (y_1',y_2').
    \end{cases}
\end{align*}
This allows us to re-write equation~\eqref{eq:first-step} as
\begin{align*}
    &\text{Var}(\sum_{y_1,y_2} f(y_1,y_2,x) S_{-2}(y_1,y_2|X=x)) \\
    &= \kappa_{-2} (\sum_{y_1,y_2} f(y_1,y_2,x)^2 P_{-2}(y_1,y_2|x) - (\sum_{y_1,y_2} f(y_1,y_2,x) P_{-2}(y_1,y_2|x))^2     ) \\
    &= \kappa_{-2} \text{Var}_{-2}(f(Y_1,Y_2,X)|X=x).
\end{align*}
On the left-hand side, the variance is over the randomness in the shift $S_t$, where on the right-hand side the variance is computed under $P_{-2}$.
Since $S_{-2}$ has mean zero, for any function $f$ with $\mathbb{E}_{-2}[f(Y_1,Y_2,X)|X=x] = 0$ we have 
\begin{align}\label{eq:S2}
\begin{split}
   &\mathbb{E}[ (\sum_{y_1,y_2} f(y_1,y_2,x) S_{-2}(y_1,y_2|x))^2] \\
   &= \kappa_{-2} \text{Var}_{-2}(f(Y_1,Y_2,X)|X=x) \\
   &= \kappa_{-2} \mathbb{E}_{-2}[f(Y_1,Y_2,X)^2|X=x].
\end{split}
\end{align}
By an analogous argument, for any function $f$ with $\mathbb{E}_{-2}[f(Y_1,Y_2,X)|X=x] = 0$ we have 
\begin{equation}\label{eq:S1S2}
\begin{split}
    &\mathbb{E}[ (\sum_{y_1,y_2} f(y_1,y_2,x) S_{-1}(y_1,y_2|x))^2] \\
    &= \kappa_{-1} (1 - \kappa_{-2}) \mathbb{E}_{-2}[f(Y_1,Y_2,X)^2|X=x].
\end{split}
\end{equation}
(Notice that $\kappa_{-2}$ is naturally upper bounded by 1: For an event with $P_{t}(\bullet|x) = 1/2$, the right-hand side of Assumption~\ref{ass:scaling} is $\kappa_t/4$. The left-hand side of $\ref{ass:scaling}$ is upper bounded by $1/4$, since $P_t + S_t$ is a  $[0,1]$-valued random variable, and the maximum variance of a $[0,1]$-valued random variable is achieved by $\text{Ber}(1/2)$. Thus, Assumption~\ref{ass:scaling} implies $\frac{1}{4} \ge \frac{\kappa_t}{4}$. Thus, we have $0 \le \kappa_{t} \le 1$.)

We now have all auxiliary results in place to investigate the MSE of $\tau^C$. For every fixed $x$ we have
\begin{align*}
  \tau(x)- \tau^C(x) &= \sum_{y_2,y_1} y_2 ( P_{-2}(y_1,y_2|x) + S_{-1}(y_1,y_2|x) + S_{-2}(y_1,y_2|x)) \\
  &- \sum_{y_2,y_1} \mathbb{E}_{-2}[Y_2|Y_1=y_1,X=x] ( P_{-2}(y_1,y_2|x) +  S_{-2}(y_1,y_2|x)) \\
  &=  \sum_{y_2,y_1} y_2 ( S_{-1}(y_1,y_2|x) + S_{-2}(y_1,y_2|x)) \\
  &- \sum_{y_2,y_1} \mathbb{E}_{-2}[Y_2|Y_1=y_1,X=x]   S_{-2}(y_1,y_2|x) ]\\
  &= \sum_{y_2,y_1} (y_2 - \mathbb{E}_{-2}[Y_2|Y_1=y_1,X=x])  S_{-2}(y_1,y_2|x) \\
  &+ \sum_{y_2,y_1} y_2 S_{-1}(y_1,y_2|x)
\end{align*}
The first inequality follows by definition. The second inequality follows from the fact that $\sum_{y_2,y_1} y_2 P_{-2}(y_1,y_2|x)=\mathbb{E}_{-2}[Y_2|X=x]$ and $\sum_{y_2,y_1} \mathbb{E}_{-2}[Y_2|Y_1=y_1,X=x]P_{-2}(y_1,y_2|x)= \mathbb{E}_{-2}[\mathbb{E}_{-2}[Y_2|Y_1,X=x]|X=x] = \mathbb{E}_{-2}[Y_2|X=x]$ and thus these terms cancel out. The third inequality follows by rearranging terms.

Since $\sum_{y_2,y_1} S_{-1}(y_1,y_2|x) = 0$, we have   $\sum_{y_2,y_1} \mathbb{E}_{-2}[Y_2|X=x] S_{-1}(y_1,y_2|x) = 0$. Combining this with the equation above, 
\begin{align*}
 &\tau(x)- \tau^C(x) \\
 &= \sum_{y_2,y_1} (y_2  - \mathbb{E}_{-2}[Y_2|Y_1=y_1,X=x]) S_{-2}(y_1,y_2|x)   \\
 &+  \sum_{y_2,y_1} (y_2  - \mathbb{E}_{-2}[Y_2|X=x]) S_{-1}(y_1,y_2|x).
\end{align*}
Squaring and taking expectations, using equation~\eqref{eq:first}, equation~\eqref{eq:S1S2}, and equation~\eqref{eq:S2} yields
\begin{align*}
    \mathbb{E}[(\tau(x)- \tau^C(x))^2|X=x] &= \kappa_{-2} \mathbb{E}_{-2}[ (Y_2  - \mathbb{E}_{-2}[Y_2|Y_1,X])^2 |X=x]  \\
    &+ \kappa_{-1} \mathbb{E}_{-2}[ (Y_2  - \mathbb{E}_{-2}[Y_2|X])^2 |X=x] + O(\kappa_{-1} \kappa_{-2}) 
\end{align*}
Now taking the expectation over $X$ yields
\begin{align*}
 \mathbb{E}[(\tau(X)- \tau^C(X))^2] &= \kappa_{-2} \mathbb{E}_{-2}[ (Y_2  - \mathbb{E}_{-2}[Y_2|Y_1,X])^2]  \\
    &+ \kappa_{-1} \mathbb{E}_{-2}[ (Y_2  - \mathbb{E}_{-2}[Y_2|X])^2] + O(\kappa_{-1} \kappa_{-2}).
\end{align*}
This completes the first claim.
Similarly,
\begin{align*}
  \tau(x)- \tau^A(x) &= \sum_{y_2,y_1} y_2 ( P_{-2}(y_2,y_1|x) + S_{-1}(y_2,y_1|x) + S_{-2}(y_2,y_1|x)) \\
  &- \sum_{y_2,y_1} y_2  P_{-2}(y_2,y_1|x)  \\
 & = \sum_{y_2,y_1} y_2  S_{-1}(y_2,y_1|x) + S_{-2}(y_2,y_1|x))  \\
  & = \sum_{y_2,y_1} (y_2 - \mathbb{E}_{-2}[Y_2|X=x]) ( S_{-1}(y_2,y_1|x) + S_{-2}(y_2,y_1|x))  
\end{align*}
 Analogously as above, squaring and taking expectations (using equation equation~\eqref{eq:first}, equation~\eqref{eq:S1S2}, and equation~\eqref{eq:S2}) yields the claim. Lastly,
\begin{align*}
   \tau(x)-  \tau^B(x)  &= \mathbb{E}_{-2}[Y_2|X=x] \\
   &+  \sum_{y_2,y_1} (y_2 - \mathbb{E}_{-2}[Y_2|X=x]) (S_{-1}(y_1,y_2|x) + S_{-2}(y_1,y_2|x))  \\
   & - \mathbb{E}_{-2}[Y_1|X=x]   \\
   &- \sum_{y_2,y_1} ( y_1 - \mathbb{E}_{-2}[Y_1|X=x] ) S_{-2}(y_1,y_2|x)  \\
   &= \mathbb{E}_{-2}[Y_2|X=x]  - \mathbb{E}_{-2}[Y_1|X=x] \\
   &+ \sum_{y_2,y_1} (y_2 - y_1 - \mathbb{E}_{-2}[Y_2 - Y_1|X=x]) S_{-2}(y_1,y_2|x) \\
   &+  \sum_{y_2,y_1} (y_2 - \mathbb{E}_{-2}[Y_2|X=x]) S_{-1}(y_1,y_2|x). 
\end{align*}
As before, squaring and taking expectations yields the claim.
\end{proof}

\subsection{Proof of Theorem~\ref{theorem:asymmetric}}

\begin{proof}
For notational simplicity, without loss of generality, we will assume that $Y_1$ is already transformed, that is $Y_1 = \tilde Y_1$. In the following, we will use $P_{-2}(y_2|y_1,x) = P_{-1}(y_2|y_1,x) = P_0(y_2|y_1,x)$ repeatedly. Using this assumption, $\tau(x) = \mathbb{E}_0[Y_2|X=x] = \mathbb{E}_0[\mathbb{E}_0[Y_2|Y_1,X]|X=x] = \mathbb{E}_0[\mathbb{E}_{-2}[Y_2|Y_1,X]|X=x]$. 
Using this, we have
\begin{align*}
    &\tau^{C}(x) - \tau(x)  \\
    &=  \sum_{y_1} \mathbb{E}_{-2}[Y_2|Y_1=y_1,X=x] p_{-1}(y_1|x) -  \sum_{y_1} \mathbb{E}_{0}[Y_2 |Y_1 = y_1,X=x] p_0(y_1|x)) \\
    &= \sum_{y_1} \mathbb{E}_{-2}[Y_2|Y_1=y_1,X=x] p_{-1}(y_1|x) -  \sum_{y_1} \mathbb{E}_{-2}[Y_2 |Y_1 = y_1,X=x] p_0(y_1|x)) \\
    &=  \sum_{y_1} \sum_{y_2} \mathbb{E}_{-2}[Y_2|Y_1=y_1,X=x] (p_{-1}(y_1,y_2|x) - p_0(y_1,y_2|x)) \\
    &= \sum_{y_1} \sum_{y_2} \mathbb{E}_{-2}[Y_2|Y_1=y_1,X=x] ( -  S_{-1}(y_1,y_2|x))
\end{align*}
Since $\mathbb{E}[S_{-1}] = 0$,
\begin{equation*}
 \mathbb{E}[(\tau^{C}(x) - \tau(x) )^2] = \text{Var}( \sum_{y_2} \sum_{y_1} \mathbb{E}_{-2}[Y_2|Y_1=y_1,X=x]   S_{-1}(y_1,y_2|x))
\end{equation*}  
Here, the outer expectation and variance are over the randomness in the shifts $S_{-1}$ and $S_{-2}$.
Now let us turn to $\tau^A(x)$.
Similarly as before,
\begin{align*}
    &\tau^{A}(x) - \tau(x)  \\
    &=  \sum_{y_1} \mathbb{E}_{-2}[Y_2|Y_1=y_1,X=x] p_{-2}(y_1|x) -  \sum_{y_1} \mathbb{E}_{0}[Y_2 |Y_1 = y_1,X=x] p_0(y_1|x)) \\
    &=  \sum_{y_1} \mathbb{E}_{-2}[Y_2|Y_1=y_1,X=x] p_{-2}(y_1|x) -  \sum_{y_1} \mathbb{E}_{-2}[Y_2 |Y_1 = y_1,X=x] p_0(y_1|x)) \\
    &= \sum_{y_1} \sum_{y_2} \mathbb{E}_{-2}[Y_2|Y_1=y_1,X=x] (p_{-2}(y_1,y_2|x) -  p_0(y_1,y_2|x))) \\
    &= \sum_{y_1} \sum_{y_2} \mathbb{E}_{-2}[Y_2|Y_1=y_1,X=x] ( - S_{-1}(y_1,y_2|x) - S_{-2}(y_1,y_2|x)) 
\end{align*}
Since $\mathbb{E}[S_{-1}] = \mathbb{E}[S_{-2}] = 0$, this term has mean zero. Using $\mathbb{E}[S_{-1}|S_{-2}] = 0$, 
\begin{align*}
   \mathbb{E}[(\tau^{A}(x) - \tau(x) )^2] &=  \text{Var}( \sum_{y_2} \sum_{y_1} \mathbb{E}_{-2}[Y_2|Y_1=y_1,X=x]  S_{-1}(y_1,y_2|x ))\\
   &+ \text{Var}( \sum_{y_2} \sum_{y_1} \mathbb{E}_{-2}[Y_2|Y_1=y_1,X=x]   S_{-2}(y_1,y_2|x)) \\
   &\ge  \mathbb{E}[(\tau^{C}(x) - \tau(x) )^2]
\end{align*}
Again, the outer expectations and variances are over the randomness in $S_{-1}$ and $S_{-2}$. Now let's turn to $\tau^B(x)$. By an analogous argument as above, we get
\begin{align*}
    \tau^{B}(x) - \tau(x) &=  \sum_{y_1} \sum_{y_2}  y_1 (p_{-2}(y_1,y_2|x) + S_{-2}(y_1,y_2|x)) \\
    &- \sum_{y_2} \sum_{y_1} \mathbb{E}_{-2}[Y_2|Y_1=y_1,X=x] (p_{-2}(y_1,y_2|x) + S_{-1}(y_1,y_2|x) + S_{-2}(y_1,y_2|x) )  \\
    &=  \sum_{y_1} \sum_{y_2} (y_1 - \mathbb{E}_{-2}[Y_2|Y_1=y_1,X=x] ) (p_{-2}(y_1,y_2|x) + S_{-2}(y_1,y_2|x))   \\
    &-  \sum_{y_2} \sum_{y_1} \mathbb{E}[Y_2|Y_1=y_1,X=x]  S_{-1}(y_1,y_2|x)
\end{align*}
Since the mean of $S_{-1}$ is zero and $S_{-2}$ and $S_{-1}$ are uncorrelated,
\begin{align*}
    \mathbb{E}[(\tau^{A}(x) - \tau(x))^2] &\ge \text{Var}(\sum_{y_2} \sum_{y_1} \mathbb{E}[Y_2|Y_1=y_1,X=x]   S_{-1}(y_1,y_2|x)). \\
    &=  \mathbb{E}[(\tau^{C}(x) - \tau(x) )^2].
\end{align*}
Again, the outer expectations and variances are over the randomness in $S_{-1}$ and $S_{-2}$. This concludes the proof.

\end{proof}

\subsection{Proof of Example~\ref{ex:noproxynoshift}}
\begin{proof}
For the second statement, please note that due to the law of large numbers $\hat\tau^B(x) \rightarrow_P \mathbb{E}_{-2} [Y_1 | X=x]$. Since by assumption $\mathbb{E}_{-2}[Y_2|X=x] \not \equiv \beta_1 \mathbb{E}_{-2} [Y_1 | X=x] + \beta_0$, $\beta_1 \hat \tau^B(x) +\beta_0 - \tau(x) \not \rightarrow_P 0$. This proves the second statement.

Let $I_{-2}$ denote the set of indices from time period $-2$ and $I_{-1}$ denote the set of indices from time period $-1$. A Taylor expansion reveals that
\begin{align*}
    \hat \tau^A(x) - \tau(x) &= \frac{1}{ \# I_{-2,x}} \sum_{i \in I_{-2,x}} Y_{2,i} - \mathbb{E}_{-2}[Y_2|X=x]  \\
    &=  \frac{1}{n_{-2}} \sum_{i \in I_{-2}} \frac{1}{ \frac{\# I_{-2,x}}{ n_{-2}}} 1_{X_i=x} (Y_{2,i} - \mathbb{E}_{-2}[Y_2|X=x])  \\
    &=  \frac{1}{n_{-2}} \sum_{ i \in I_{-2} } \frac{1_{X_i = x}}{\mathbb{P}_{-2}(X=x)}  ( Y_{2,i} - \mathbb{E}_{-2}[Y_2 |X=x])  + o_P(1/\sqrt{n}).
\end{align*}
Thus,
\begin{align*}
    \sqrt{n_{-2}} ( \hat \tau^A(x) - \tau(x)) \rightharpoonup \mathcal{N}(0, \frac{1}{P_{-2}(X=x)}\text{Var}_{-2}(Y_2|X=x) )
\end{align*}
As $n_{-1}/n_{-2} \rightarrow \rho$ and $n = n_{-1} + n_{-2}$,
\begin{equation*}
    \sqrt{n} ( \hat \tau^A(x) - \tau(x)) \rightharpoonup \mathcal{N}(0, ( 1+ \rho) \frac{1}{P_{-2}(X=x)}\text{Var}_{-2}(Y_2|X=x) ).
\end{equation*}

For $\tau^C$ the proof proceeds analogously but is a bit more technical:
\begin{align*}
    &\hat \tau^C(x) - \tau(x))  \\
    &= \frac{1}{ \# I_{-1,x}} \sum_{i \in I_{-1,x}} \frac{1}{ \# I_{-2,x,Y_{1,i}}} \sum_{i' \in I_{-2,x,Y_{1,i}}} Y_{2,i'} - \mathbb{E}_{-2}[Y_2|X=x] \\
      & =  \frac{1}{ n_{-1}} \sum_{i \in I_{-1}} \frac{ 1_{X_i=x_i } }{ \# I_{-1,x} /n_{-1}} \frac{1}{ n_{-2}} \sum_{i' \in I_{-2}}  \frac{1_{X_{i'}=x_{i'}  , Y_{1,i'} = Y_{1,i}}}{ \# I_{-2,x,Y_{1,i}} /n_{-2}} (Y_{2,i'} - \mathbb{E}_{-2}[Y_2|X=x]) \\
      &= \frac{1}{ n_{-1}} \sum_{i \in I_{-1}} \frac{ 1_{X_i=x_i } }{ \# I_{-1,x} /n_{-1}} \frac{1}{ n_{-2}} \sum_{i' \in I_{-2}}  \frac{1_{X_{i'}=x_{i'}  , Y_{1,i'} = Y_{1,i}}}{ \# I_{-2,x,Y_{1,i}} /n_{-2}} \\
      & \qquad \qquad \cdot (\mathbb{E}_{-2}[Y_2|Y_1 = Y_{1,i}, X=x] - \mathbb{E}_{-2}[Y_2|X=x]) \\
      &+ \frac{1}{ n_{-1}} \sum_{i \in I_{-1}} \frac{ 1_{X_i=x_i } }{ \# I_{-1,x} /n_{-1}} \frac{1}{ n_{-2}} \sum_{i' \in I_{-2}}  \frac{1_{X_{i'}=x_{i'}  , Y_{1,i'} = Y_{1,i}}}{ \# I_{-2,x,Y_{1,i}} /n_{-2}} (Y_{2,i'} - \mathbb{E}_{-2}[Y_2|Y_1 = Y_{1,i}, X=x]) \\
\end{align*}
The first part can be drastically simplified and we can add an additional sum over $y$ to decouple the second term:
\begin{align*}
&\hat \tau^C(x) - \tau(x)) \\
     &=\frac{1}{ n_{-1}} \sum_{i \in I_{-1}} \frac{ 1_{X_i=x } }{ \# I_{-1,x} /n_{-1}}(\mathbb{E}_{-2}[Y_2|Y_1 = Y_{1,i}, X=x] - \mathbb{E}_{-2}[Y_2|X=x]) \\
      &+ \frac{1}{ n_{-1}} \sum_{i \in I_{-1}} \frac{ 1_{X_i=x } }{ \# I_{-1,x} /n_{-1}} \frac{1}{ n_{-2}} \sum_{i' \in I_{-2}}  \frac{1_{X_{i'}=x_{i'}  , Y_{1,i'} = Y_{1,i}}}{ \# I_{-2,x,Y_{1,i}} /n_{-2}} (Y_{2,i'} - \mathbb{E}_{-2}[Y_2|Y_1 = Y_{1,i}, X=x]) \\
       &= \frac{1}{ n_{-1}} \sum_{i \in I_{-1}} \frac{ 1_{X_i=x } }{\# I_{-1,x} /n_{-1}}(\mathbb{E}_{-2}[Y_2|Y_1 = Y_{1,i}, X=x] - \mathbb{E}_{-2}[Y_2|X=x]) \\
       &+ \sum_y \frac{1}{ n_{-1}} \sum_{i \in I_{-1}} \frac{ 1_{X_i=x ,y_i=y } }{ \# I_{-1,x} /n_{-1}} \frac{1}{ n_{-2}} \sum_{i' \in I_{-2}}  \frac{1_{X_{i'}=x_{i'}  , Y_{1,i'} =  y}}{ \# I_{-2,x,y} /n_{-2}} (Y_{2,i'} - \mathbb{E}_{-2}[Y_2|Y_1 = y, X=x])
\end{align*}
Now by the law of large numbers,
\begin{align*}
    &\hat \tau^C(x) - \tau(x)) \\
    &=  \frac{1}{ n_{-1}} \sum_{i \in I_{-1}} \frac{ 1_{X_i=x } }{P_{-2}(X=x)}(\mathbb{E}_{-2}[Y_2|Y_1 = Y_{1,i}, X=x] - \mathbb{E}_{-2}[Y_2|X=x]) \\
    &+\sum_y  \frac{ P_{-2}(X=x,Y_1=y ) }{ P_{-2}(X=x)} \frac{1}{ n_{-2}} \sum_{i' \in I_{-2}}  \frac{1_{X_{i'}=x_{i'} , Y_{1,i'} = y}}{P_{-2}(X=x,Y_1=y)} (Y_{2,i} - \mathbb{E}_{-2}[Y_2|Y_1 = y, X=x]) \\
    &+ o_P(1/\sqrt{n})\\
    &=  \frac{1}{ n_{-1}} \sum_{i \in I_{-1}} \frac{ 1_{X_i=x } }{P_{-2}(X=x)}(\mathbb{E}_{-2}[Y_2|Y_1 = Y_{1,i}, X=x] - \mathbb{E}_{-2}[Y_2|X=x]) \\
    &  \frac{1}{ n_{-2}} \sum_{i' \in I_{-2}} \sum_y  \frac{1_{X_{i'}=x_{i'} , Y_{1,i'} = y}}{P_{-2}(X=x)} (Y_{2,i} - \mathbb{E}_{-2}[Y_2|Y_1 = y, X=x]) \\
    &+ o_P(1/\sqrt{n}) \\
    &=  \frac{1}{ n_{-1}} \sum_{i \in I_{-1}} \frac{ 1_{X_i=x } }{P_{-2}(X=x)}(\mathbb{E}_{-2}[Y_2|Y_1 = Y_{1,i}, X=x] - \mathbb{E}_{-2}[Y_2|X=x]) \\
    &  \frac{1}{ n_{-2}} \sum_{i' \in I_{-2}}  \frac{1_{X_{i'}=x_{i'} }}{P_{-2}(X=x)} (Y_{2,i'} - \mathbb{E}_{-2}[Y_2|Y_1 = Y_{1,i'}, X=x]) \\
    &+ o_P(1/\sqrt{n}) 
\end{align*}
Thus, by the CLT, with $\frac{n_{-1}}{n_{-2}} \rightarrow \rho$ we have
\begin{equation*}
    \sqrt{n}( \hat \tau^C(x) - \tau(x)) \rightharpoonup \mathcal{N}(0,\sigma_C^2),
\end{equation*}
where
\begin{align*}
    \sigma_C^2 &= \frac{1 + \rho}{\rho } \frac{1}{P_{-2}(X=x)}\text{Var}_{-2}(\mathbb{E}_{-2}[Y_2|Y_1,X=x]|X=x) \\
    &+ ( 1+ \rho) \frac{1}{P_{-2}(X=x)}\text{Var}_{-2}(Y_2 - \mathbb{E}_{-2}[Y_2|Y_1,X=x]|X=x).
\end{align*}
Thus, $\tau_A^2 < \tau_C^2$ if and only if $\rho < 1$.

\end{proof}

\subsection{Proof of Example~\ref{ex:proxy}}

\begin{proof}
Please note that following the proof of Example~\ref{ex:noproxynoshift}, the estimators $\hat \tau^A(x)$, $\hat \tau_{\text{scaled}}^B(x) := \beta_1 \tau^B(x) + \beta_0$ and $\hat \tau^C(x)$ are all asymptotically unbiased. Thus, in the following we will compare their respective asymptotic variances. For $\hat \tau_\text{scaled}^B(x)$ we have asymptotic variance
\begin{align*}
    &\frac{1 + \rho}{\rho P_{-2}(X=x)} \text{Var}(\beta_1 Y_1 + \beta_0|X=x)  \\
    &=     \frac{1 + \rho}{\rho P_{-2}(X=x)} \text{Var}(\mathbb{E}_{-2}[Y_2|Y_1,X=x]|X=x)  \\
\end{align*}
From the proof of Example~\ref{ex:noproxynoshift} we know that the asymptotic variance of $\hat \tau^A(x)$ is
\begin{align*}
    & \frac{1 + \rho}{ P_{-2}(X=x)} \text{Var}(Y_2 - \mathbb{E}_{-2}[Y_2|Y_1,X=x]|X=x) \\
    &+ \frac{1 + \rho}{ P_{-2}(X=x)} \text{Var}(\mathbb{E}_{-2}[Y_2|Y_1,X=x]|X=x).
\end{align*}
Similarly, for $\hat \tau^C$ we know that the asymptotic variance is
\begin{align*}
    & (1 + \rho) \frac{1}{ P_{-2}(X=x)} \text{Var}(Y_2 - \mathbb{E}_{-2}[Y_2|Y_1,X=x]|X=x) \\
    &+ \frac{1 + \rho }{\rho } \frac{1}{ P_{-2}(X=x)} \text{Var}(\mathbb{E}_{-2}[Y_2|Y_1,X=x]|X=x).
\end{align*} 
Since $ \rho > 1$, and $\mathbb{E}_{-2}[Y_2|Y_1,X] = \beta_1 Y_1 + \beta_0$ the claim follows.
\end{proof}

\end{document}